\documentclass[preprint,review,12pt,a4paper]{elsarticle}

\usepackage{amsmath}
\usepackage{amssymb}
\usepackage{graphicx}
\usepackage{geometry}
\usepackage{hyperref}
\usepackage{xspace}
\usepackage{booktabs}
\usepackage{tabularx}
\usepackage{array}
\usepackage{pdflscape}   
\usepackage{booktabs}

\usepackage{parskip}
\begin{document}

\title{Reinforcement Learning with a Bilevel World-Model Architecture for Scan-Order Optimisation in Laser Directed Energy Deposition}



\author[1]{Xian Wu}
\author[2]{Haoran Li}
\author[1]{Yuanqi Chu}
\author[2]{Dongbin Zhao}\corref{cor2}
\ead{dongbin.zhao@ia.ac.cn}
\author[1]{Bin Wang}\corref{cor1}
\ead{Bin.Wang@brunel.ac.uk}

\affiliation[1]{
  organization={College of Engineering, Design and Physical Sciences, Brunel University London},
  city={Uxbridge},
  postcode={UB8 3PH},
  country={United Kingdom}
}

\affiliation[2]{
  organization={The State Key Laboratory of Management and Control for Complex Systems, Institute of Automation, Chinese Academy of Sciences},
  city={Beijing},
  postcode={100190},
  country={China}
}


\cortext[cor1]{Corresponding author.}
\cortext[cor2]{Corresponding author.}


\begin{abstract}
  Scan-order design in laser directed energy deposition (LDED) is a delayed, path-dependent thermo-mechanical decision problem whose quality becomes apparent only after the complete deposition and cooling cycle. Existing scan heuristics and scalarised optimisation methods can evaluate completed sequences, but they lack a native mechanism for translating terminal finite-element responses into track-by-track ordering decisions. This work formulates LDED scan-order optimisation as a finite-horizon, permutation-constrained reinforcement-learning problem and addresses it through a bilevel finite-element-teacher-labelled AI workflow. In this workflow, a surrogate-assisted teacher-guided optimisation loop learns the Abaqus-labelled response landscape and provides a computationally tractable terminal-reward environment for policy training. A frozen Maskable Proximal Policy Optimization (MaskablePPO) policy is then used to generate legal scan-order candidates, which are subsequently evaluated through independent Abaqus thermo-mechanical simulations. The teacher-validated rankings demonstrate bounded, \(N\)-dependent PPO policy-generation value rather than record-level dominance over the mature surrogate-assisted optimiser. The strongest scan orders in the current evidence pool are obtained by the teacher-guided surrogate-assisted optimisation loop, whereas the PPO-generated candidates demonstrate that a standard policy-gradient agent can autonomously generate physically admissible scan orders that reach competitive regions of the native response landscape, with the strongest rank concentration in smaller track-count settings and a clear reliability boundary as the sequence horizon increases. Beyond ranking performance, the teacher-labelled response landscape reveals a physically gated, lexicographic reward hierarchy in which warpage admissibility constitutes the primary constraint, plastic-strain response serves as a safety filter and residual-stress-related improvement is applied conditionally within the admissible region. The validated sequences further suggest an interpretable scale-separated ordering tendency, combining global spatial dispersion with local structured grouping rather than reproducing a single predefined scan rule. These results establish a computational pathway from predefined scan-strategy selection toward finite-element-teacher-validated policy generation, enabling the proposal and physical interpretation of scan-order structures beyond a finite repertoire of existing engineering rules while preserving the need for independent finite-element teacher validation.
\end{abstract}


\maketitle


\section{Introduction}


Laser directed energy deposition (LDED) is inherently sequential: material is deposited track by track through localised laser--material interaction, and each new track modifies the thermal and mechanical state inherited by subsequent tracks \cite{Caiazzo2022}. In a multi-track build, the scan order therefore governs not only the geometric progression of deposition, but also the return time between neighbouring tracks, the accumulation and redistribution of heat, the cooling sequence and the evolution of thermo-mechanical history. As a result, macroscopic distortion, equivalent plastic strain and residual-stress-related response emerge after deposition and cooling from the accumulated sequence history, rather than from any isolated local decision \cite{Ding2024, Denlinger2015}. Scan-order design should therefore be treated as a delayed, path-dependent thermo-mechanical decision problem, in which each local ordering choice reshapes the physical context for the remaining deposition sequence.


Existing scan-order design approaches do not naturally operate at this decision level. Conventional strategies, including raster, alternating, island-style, centre-out and dispersion-based scanning, encode valuable engineering intuition, but they prescribe complete ordering patterns before the thermo-mechanical consequence of each sequence is known \cite{Dar2025}. Optimisation-oriented studies can compare candidate scan orders using scalarised distortion, plastic-strain or residual-stress-related objectives, but they typically evaluate completed sequences rather than learn the sequential rule by which each track should be selected \cite{He2025,Zhao2025}. Recent reinforcement-learning studies have begun to explore laser scan-path planning, particularly in powder-bed systems or thermal-control settings, yet finite-element-teacher-labelled LDED track-order learning under hierarchical mechanical constraints remains insufficiently developed \cite{Qin2024,Dou2026}. The central gap is therefore not simply the size of the combinatorial search space, but the absence of a mechanism that translates delayed finite-element responses into track-by-track physical decision rules.


This gap motivates a reinforcement-learning formulation of LDED scan-order optimisation. Rather than treating scan order as a completed permutation to be ranked after the fact, the problem is recast as a finite-horizon, permutation-constrained sequential decision process, consistent with recent reinforcement-learning formulations for additive-manufacturing scan-path and toolpath generation \cite{Qin2024,Dou2026}. At each decision step, the state encodes the partially completed deposition sequence together with the set of remaining admissible tracks, while the action consists of selecting one previously unvisited track. Each action extends the sequence, removes the chosen track from the legal action set and alters the thermo-mechanical context inherited by subsequent decisions. The episode terminates only after all tracks have been selected, at which point the completed scan-order permutation receives a terminal reward derived from its post-cooling thermo-mechanical response. This formulation establishes a clean decision interface: the policy is responsible for generating valid track-by-track ordering rules, while the physical evaluation of each completed sequence can be supplied independently, either by a computationally efficient surrogate during policy training or by direct finite-element teacher validation during final assessment \cite{Kim2026,Francon2020}.


The principal bottleneck in this reinforcement-learning formulation lies not in generating legal scan-order permutations, but in acquiring physically meaningful rewards. Because the manufacturing quality of a scan order is determined only after the complete deposition and cooling cycle, the reward cannot be derived from geometric or local sequence features alone. It must instead be grounded in terminal thermo-mechanical quantities, including the \(U_2\)-based vertical-displacement response for warpage-related distortion, \(\mathrm{PEEQ}\) for accumulated plastic deformation and \(\mathrm{Surface}_T\) for residual-stress-related response. These quantities are not interchangeable scalar objectives. They form a physically ordered hierarchy in which warpage-related geometric admissibility defines the primary constraint, plastic-strain response serves as a subsequent safety filter and residual-stress-related improvement is meaningful only within the admissible response region \cite{Xue2025}. Consequently, the reward signal is terminal, computationally expensive and hierarchically constrained. This necessitates a framework that decouples policy generation from physical evaluation: a surrogate approximates terminal rewards during policy training, while direct finite-element teacher validation is reserved for final performance assessment \cite{Kim2026,Francon2020}.


This study addresses the above requirements through a bilevel finite-element-teacher-labelled AI workflow that decouples teacher-guided response-landscape optimisation from policy-gradient scan-order generation. A native multi-\(N\) set of Abaqus-labelled scan-order cases is first assembled across multiple track-count settings. From these teacher-labelled cases, a supervised surrogate terminal-reward model is trained to learn the mapping from completed scan-order descriptors to Abaqus-derived responses and reward. This surrogate-assisted teacher-guided optimisation loop provides the strongest current search mechanism within the teacher-labelled response landscape and supplies a computationally efficient terminal-reward environment for policy optimisation. Within this learned reward environment, a mask-constrained MaskablePPO policy learns to generate valid scan-order permutations by selecting exclusively from the remaining admissible tracks at each decision step \cite{Schulman2017,Raffin2021,Huang2022}. After training, the policy checkpoint is frozen and used solely for candidate generation. Final admissibility, ranking and physical interpretation are then obtained through independent Abaqus thermo-mechanical simulations and output-database metric extraction. This separation makes policy optimisation computationally tractable while ensuring that performance claims, competitiveness assessment and scan-order interpretation remain grounded in direct finite-element teacher validation \cite{Kim2026,Francon2020}.


This study contributes a bilevel AI framework for LDED scan-order optimisation by combining supervised FEA-teacher surrogate modelling with policy-gradient reinforcement learning. The surrogate-assisted teacher-guided optimisation loop identifies high-quality candidates and learns the Abaqus-labelled response landscape defined by \(U_2\), \(\mathrm{PEEQ}\), \(\mathrm{Surface}_T\) and the derived reward. The MaskablePPO layer then distils this learned terminal-reward landscape into an executable sequential decision policy that can autonomously generate legal scan-order candidates. The resulting PPO-generated candidates are not accepted on the basis of surrogate scores alone; instead, they are independently evaluated using Abaqus teacher simulations. The supported PPO claim is therefore bounded: the policy demonstrates legal scan-order generation and teacher-validated small-\(N\)/top-\(k\) competitiveness, but it is not claimed to outperform the mature surrogate-assisted optimiser, establish global optimality or solve arbitrary-\(N\) scan-order optimisation.

Beyond the policy-generation evidence, the teacher-labelled response landscape demonstrates that scan-order quality is more appropriately represented by a physically gated reward hierarchy than by flat scalarisation of distortion, plasticity and stress-related metrics, with warpage admissibility as the primary constraint, plastic-strain response as a safety filter and residual-stress-related improvement applied conditionally within the admissible region \cite{Xue2025}. The validated candidates further suggest interpretable ordering tendencies beyond conventional human-designed heuristics, combining global spatial dispersion with local structured grouping \cite{Dar2025,Ding2024}. Together, these results establish a computational pathway from predefined scan-strategy comparison toward finite-element-teacher-validated policy generation, enabling the proposal, validation and physical interpretation of scan-order structures beyond a finite repertoire of existing engineering rules while preserving the requirement that physical claims remain anchored to independent finite-element teacher validation.

\section{Methods}






\subsection{Multi-N benchmark suite and finite-element teacher-labelled response set}

The scan-order optimisation task is formulated as a controlled multi-N LDED benchmark. For each track-count setting, the substrate geometry, SS316L material definition, boundary conditions and process parameters are held fixed, so that the only variable under investigation is the activation order of the deposition tracks. This controlled design isolates the thermo-mechanical consequences of scan-order sequencing from other process variables and enables direct comparison of candidate permutations under a consistent finite-element teacher.

The primary benchmark comprises four native track-count settings: N12, N16, N24 and N40. These settings are used to evaluate whether the proposed framework can generate and compare scan-order permutations across varying sequence lengths while preserving within-\(N\) physical comparability. An auxiliary fixed-N32 teacher-labelled reference set is retained for baseline comparison and legacy fixed-\(N\) learned-strategy analysis. The N32 set is reported where relevant, but the primary native multi-\(N\) claims of this study are grounded in the N12, N16, N24 and N40 settings.

\begin{figure}[tb]
  \centering
  \includegraphics[width=1\textwidth]{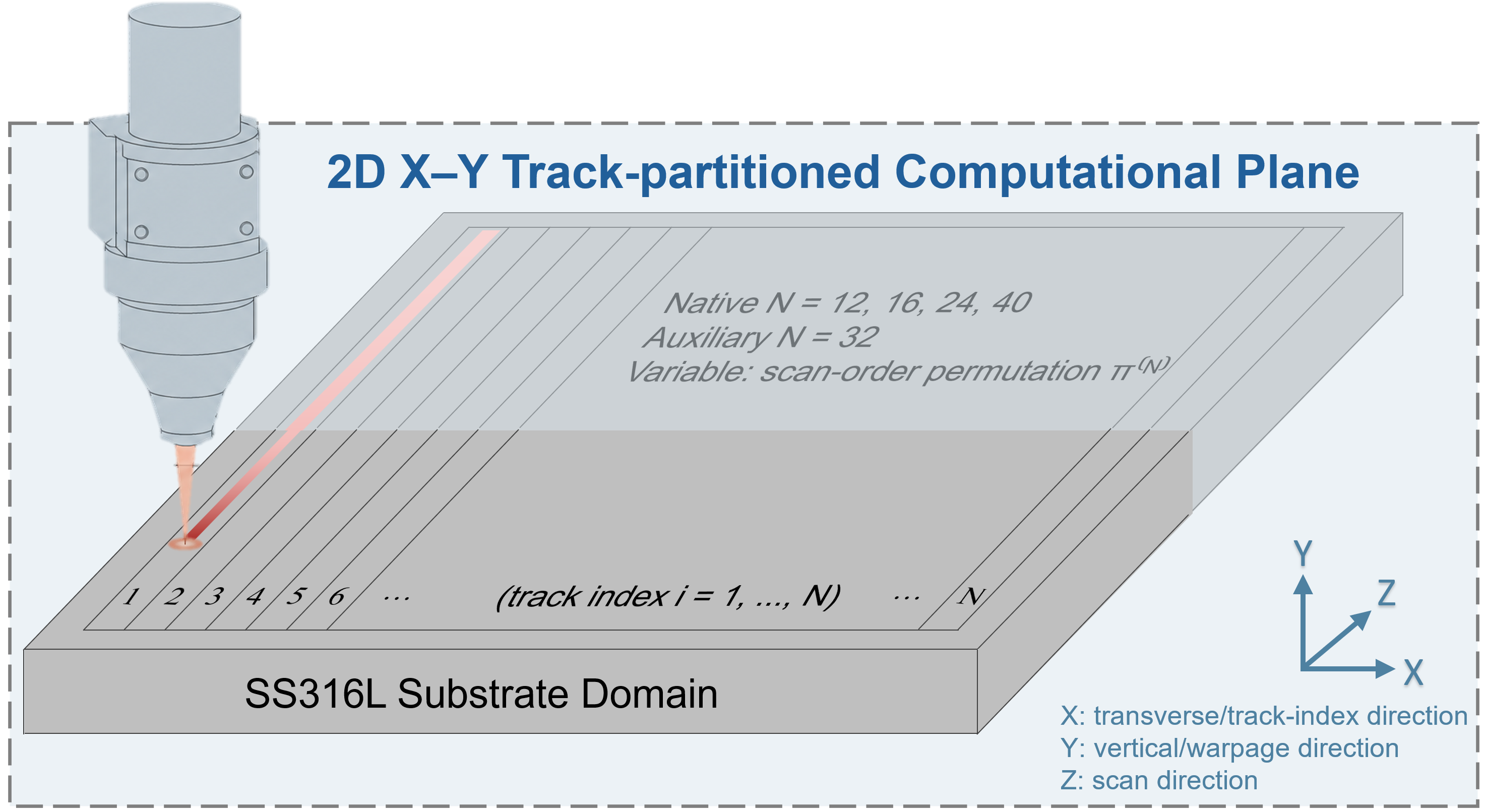}
  \caption{Multi-\(N\) LDED benchmark geometry and two-dimensional track-partitioning convention. The grey block denotes the SS316L substrate domain, while the blue shaded plane schematically represents the two-dimensional track-partitioned computational section used to define scan-order permutations in the finite-element benchmark. For each \(N\)-track setting, the track set is defined as \(\mathcal{T}^{(N)} = \{1, \dots, N\}\), and a candidate scan order \(\pi^{(N)}\) is a complete permutation of this set. \(N = 12, 16, 24\) and \(40\) constitute the native multi-\(N\) benchmark suite, while \(N = 32\) is retained as an auxiliary fixed-track-count reference.}
  \label{fig:multi-n-benchmark-track-partitioning}
\end{figure}

For an N-track setting, the track set is denoted as
\[
  \mathcal{T}^{(N)} = \{1, 2, \ldots, N\}.
\]
A candidate scan order is a complete permutation \(\sigma^{(N)}\) of this track set, where each track appears exactly once. The admissible scan-order space, denoted as \(\Sigma^{(N)}\), therefore contains all \(N!\) possible permutations. Because the absolute magnitude of displacement, plastic strain and stress-related responses varies with track count and geometry, all metric normalisation, feasibility assessment and policy ranking are conducted within each N-track setting unless cross-N comparison is explicitly stated.

The benchmark is organised around a finite-element teacher-labelled response set, denoted as \(\mathcal{D}\). Each admitted instance in \(\mathcal{D}\) associates a completed scan-order permutation with its terminal thermo-mechanical response obtained from Abaqus analysis. The response vector includes the principal physical quantities used for reward construction, feasibility assessment and policy evaluation: vertical displacement range, denoted as \(\Delta U_2\), for warpage-related distortion; equivalent plastic strain, denoted as \(\mathrm{PEEQ}\), for irreversible deformation accumulation; and surface tensile-stress index, denoted as \(\mathrm{Surface}_T\), for residual-stress-related response. Additional stress measures are retained as diagnostic quantities.

To ensure traceability and physical credibility, provenance and validity metadata are maintained separately from the physical response instance. The provenance label records the candidate origin, including engineering baselines, predefined heuristic strategies, learned policies, surrogate-guided selection, uncertainty-based selection, diversity sampling and local refinement. The validity status records whether the Abaqus analysis completed successfully, satisfied the prescribed numerical-convergence criteria and produced the required output database fields.

A candidate is admitted into the teacher-labelled response set \(\mathcal{D}\) only when its validity status confirms successful completion, numerical convergence and availability of the necessary output fields. Instances failing these quality gates are retained in the metadata record for auditability, but are excluded from teacher-labelled reward assignment and final physical ranking unless explicitly used in a separate diagnostic failure analysis. In this way, \(\mathcal{D}\) constitutes a quality-gated collection of finite-element teacher responses that serves as the physically grounded evidence base for surrogate reward modelling, PPO terminal-reward environment construction, teacher-guided candidate analysis and subsequent scan-order ranking.

\subsection{Reinforcement-learning formulation of scan-order optimisation}

LDED scan-order optimisation is formulated as a finite-horizon reinforcement-learning problem in which one episode constructs one complete deposition-track activation sequence. This formulation is appropriate because the design variable is not a continuous process parameter along a prescribed path, but the order in which spatially distributed deposition tracks are activated. The policy therefore learns to generate a legal scan-order permutation, while the physical quality of that permutation is assessed only after the completed sequence has been evaluated through the terminal thermo-mechanical response.

For a given N-track benchmark, a completed scan order is represented as a discrete permutation sequence:
\[
  \sigma^{(N)} = (\sigma_1, \sigma_2, \ldots, \sigma_N),
\]
where each element identifies one physical deposition track. The episode length is \(N\). Before decision step \(t\), the partial scan order is:
\[
  \sigma_{1:t-1}^{(N)} = (\sigma_1, \sigma_2, \ldots, \sigma_{t-1}).
\]
The legal action set at step \(t\) contains the tracks that have not yet been selected:
\[
  \mathcal{A}_t^{(N)} = \mathcal{T}^{(N)} \setminus \{\sigma_1, \sigma_2, \ldots, \sigma_{t-1}\}.
\]
The policy \(\pi_\theta\) selects the next track from this legal set:
\[
  a_t \in \mathcal{A}_t^{(N)}.
\]
The selected action is concatenated with the partial sequence:
\[
  \sigma_{1:t}^{(N)} = \sigma_{1:t-1}^{(N)} \oplus (a_t),
\]
where \(\oplus\) denotes sequence concatenation.
At the algorithmic level, this transition extends the partial permutation and updates the legal-action mask. At the physical level, it assigns one deposition track to a specific activation time in the future thermo-mechanical loading history. The action is therefore not merely a symbolic index choice; it determines the next admissible spatial location of transient heat input under the manufacturing constraint that each track can be deposited only once. A complete episode progressively defines the spatio-temporal heat-source activation schedule that is subsequently scored by the surrogate terminal-reward environment during PPO training and independently assessed through Abaqus teacher validation during final physical evaluation.

This formulation distinguishes scan-order learning from process-control reinforcement learning in additive manufacturing. The policy does not regulate laser power, scan velocity or melt-pool state during deposition. These quantities are fixed within each benchmark setting. Instead, the policy controls the combinatorial ordering of predefined deposition tracks, and its objective is to generate physically favourable scan-order permutations under delayed terminal feedback.

The permutation constraint is enforced through legal-action masking. At each decision step, an action is legal only if the corresponding track belongs to the current N-track benchmark and has not appeared in the partial scan order. Previously selected tracks are invalid because each track can be activated only once. Track indices beyond the current \(N\) are also invalid because the PPO environment uses a fixed maximum action dimension corresponding to the largest native benchmark, N40.

For track \(j\) at step \(t\), the mask value is defined as:
\[
  M_{j,t} =
  \begin{cases}
    1 & \text{if track \(j\) is legal at step \(t\)}, \\
    0 & \text{otherwise}.
  \end{cases}
\]
The zero entry covers both already selected tracks and indices outside the current \(N\)-track setting. The masked policy distribution is constrained as:
\[
  \pi_\theta(a_t = j \mid s_t, M_t) = 0 \quad \text{if } M_{j,t} = 0.
\]
For legal actions, the policy distribution is normalised only over tracks with \(M_{j,t} = 1\). This masking mechanism ensures that each generated sequence is a valid scan-order permutation \cite{Huang2022}. It also allows a single MaskablePPO environment structure to operate across N12, N16, N24 and N40 by preserving a fixed action-space interface while changing the legal-action mask according to the current track-count setting.

The action mask is therefore not a cosmetic implementation detail. It is the mechanism that translates the manufacturing legality constraint into the reinforcement-learning policy. Without masking, the policy could allocate probability to repeated tracks or to track indices that do not exist in the current benchmark. With masking, policy-gradient exploration is restricted to the admissible permutation space.

The scan-order reward is sparse and terminal. Intermediate decisions are assigned zero reward:
\[
  R_t = 0 \quad \text{for } t < N.
\]
This design reflects the path-dependent nature of LDED thermo-mechanical response. A local track choice does not uniquely determine the final post-cooling state, because its effect can be relaxed, amplified or redistributed by later thermal cycles. Warpage-related displacement, equivalent plastic strain and residual-stress-related response are accumulated outcomes of the complete deposition and cooling history. The reward is therefore assigned only after the full scan-order sequence has been constructed.

During MaskablePPO training, the terminal reward is supplied by a frozen surrogate terminal-reward environment derived from finite-element teacher-labelled scan-order data:
\[
  R_N = R_{\text{surrogate}}(\sigma^{(N)}).
\]
The construction, validation accuracy and physical role of this surrogate reward environment are described in Section 2.3. The surrogate environment provides computationally tractable terminal feedback for PPO training, but it is not treated as the physical teacher. PPO was not trained online with Abaqus in the loop. Instead, Abaqus simulations were reserved for independent teacher validation of PPO-generated candidates.

For final physical assessment, the generated scan orders are converted into Abaqus cases and evaluated using thermo-mechanical finite-element simulations. The resulting teacher-extracted metrics, rather than surrogate predictions, define the reported physical ranking and comparison against the native teacher-labelled reference pool.

The policy defines a probability distribution over legal actions conditioned on the current state and legal-action mask:
\[
  \pi_\theta(a_t \mid s_t, M_t),
\]
where \(\theta\) denotes the trainable policy parameters, \(s_t\) denotes the current scan-order construction state and \(M_t\) denotes the legal-action mask. The policy objective is to maximise the expected surrogate terminal reward over complete valid scan-order permutations:
\[
  \max_{\theta} \ \mathbb{E}\bigl[ R_{\text{surrogate}}(\sigma^{(N)}) \bigr]
\]
subject to \(\sigma^{(N)}\) being a valid permutation of the current \(N\)-track benchmark.

MaskablePPO is used because it natively supports invalid-action masking during policy optimisation \cite{Schulman2017,Raffin2021,Huang2022}. This is essential for scan-order generation, where the policy must explore a factorial permutation space without producing repeated-track sequences. In this study, MaskablePPO provides the policy-gradient mechanism for learning legal scan-order generation in the surrogate terminal-reward environment. The subsequent Abaqus teacher-validation stage determines whether the PPO-generated scan orders remain physically admissible and competitive under high-fidelity thermo-mechanical evaluation.

\begin{table}[tb]
  \centering
  \caption{Mapping between LDED scan-order optimisation and the reinforcement-learning formulation}
  \label{tab:2.2}
  \renewcommand{\arraystretch}{1.45}
  \begin{tabularx}{\textwidth}{>{\raggedright\bfseries\arraybackslash}p{3.6cm} >{\raggedright\arraybackslash}X}
    \toprule
    \textbf{RL component} & \textbf{Definition in this study} \\
    \midrule
    Episode
    & Construction of a complete scan-order permutation for a given \(N\)-track benchmark \\

    Episode length
    & \(N\) sequential decisions for an \(N\)-track setting \\

    State
    & Partial scan order, selected-track history, remaining legal tracks, current step index, previously selected track and geometric descriptors \\

    Action
    & Selection of one currently legal, unvisited deposition track \\

    Legal-action mask
    & Excludes already selected tracks and track indices outside the current \(N\)-track setting \\

    Fixed action dimension
    & Maximum native action dimension corresponding to the N40 setting, with unavailable tracks masked for smaller \(N\) \\

    Transition
    & Appending the selected track to the partial sequence and updating the legal-action mask \\

    Intermediate reward
    & Zero before sequence completion \\

    Training terminal reward
    & Surrogate-predicted terminal reward derived from finite-element-teacher-labelled data \\

    Final physical assessment
    & Independent Abaqus teacher validation of PPO-generated scan orders \\

    Policy algorithm
    & MaskablePPO with legal-action masking \\
    \bottomrule
  \end{tabularx}
\end{table}

\subsection{FEA-teacher-labelled surrogate reward environment}

The PPO training environment was constructed as a supervised terminal-reward surrogate fitted to finite-element teacher-labelled scan-order data. This surrogate was introduced to decouple policy-gradient sampling from the computational cost of full Abaqus evaluation, while retaining a reward signal calibrated from high-fidelity thermo-mechanical teacher responses \cite{Kim2026,Francon2020}. The surrogate therefore provides the learned reward environment for PPO training; it does not replace the finite-element teacher used for final physical validation.

In the bilevel AI workflow, this supervised surrogate has a role distinct from the PPO policy. The surrogate-assisted teacher-guided optimisation loop learns the Abaqus-labelled response landscape and supports candidate screening, ranking and response-landscape analysis. MaskablePPO is then trained inside the frozen surrogate terminal-reward environment to convert this learned landscape into an executable sequential decision policy. Thus, the surrogate is a supervised AI model used to make policy training computationally feasible, whereas PPO provides the policy-gradient scan-order generator.

The native teacher-labelled pool used for surrogate fitting is denoted as \(\mathcal{D}_{\text{native}}\):
\[
  \mathcal{D}_{\text{native}} = \mathcal{D}_{12} \cup \mathcal{D}_{16} \cup \mathcal{D}_{24} \cup \mathcal{D}_{40}
\]
where \(\mathcal{D}_{12}\), \(\mathcal{D}_{16}\), \(\mathcal{D}_{24}\) and \(\mathcal{D}_{40}\) are the valid Abaqus teacher-labelled subsets for N12, N16, N24 and N40, respectively. This native pool contains 552 teacher-labelled scan-order instances in total, comprising 78 cases for N12, 78 cases for N16, 190 cases for N24 and 206 cases for N40. The auxiliary fixed-N32 reference set was excluded from the primary native surrogate fitting so that the reward environment used for PPO training and native response-landscape ranking was defined only by the N12, N16, N24 and N40 teacher-labelled data.

For each scan-order instance \(i\) in an N-track setting, the Abaqus teacher returns a terminal thermo-mechanical response vector:
\[
  \mathbf{y}_i^{(N)} = \left[\Delta U_{2,i},\ \mathrm{PEEQ}_{\max,i},\ \mathrm{Surface}_{T,i},\ \mathrm{Mises}_{\max,i}\right]
\]
where \(\Delta U_2\) is the vertical displacement range used to represent warpage-related distortion, \(\mathrm{PEEQ}_{\max}\) is the maximum equivalent plastic strain, \(\mathrm{Surface}_T\) is the surface tensile-stress index and \(\mathrm{Mises}_{\max}\) is retained as a diagnostic stress measure. These quantities are terminal responses extracted after completion of the finite-element simulation and post-processing, rather than intermediate step-wise labels.

The scalar reward used for PPO training was derived from this teacher response vector through a hierarchical terminal-reward transformation:
\[
  r_i = R_{\text{lex}}(\mathbf{y}_i^{(N)})
\]
where \(R_{\text{lex}}\) denotes the \(U_2\)--\(\mathrm{PEEQ}\)--\(\mathrm{Surface}_T\) lexicographic reward operator. The implemented target, referred to in the data pipeline as \texttt{reward\_lex\_u2\_peeq\_surfacet} for traceability, follows a physically ordered hierarchy: \(U_2\)-related warpage is treated as the primary response, \(\mathrm{PEEQ}\) defines the plasticity-safety layer and \(\mathrm{Surface}_T\) provides the secondary residual-stress-related ranking signal. \(\mathrm{Mises}\) stress is retained for diagnostic assessment and is not part of the primary PPO reward target.

This reward construction is intentionally hierarchical rather than a flat weighted scalarisation. In a flat weighted sum, improvement in a secondary stress-related metric could compensate for an unfavourable warpage response. The lexicographic transformation avoids this compensation by preserving the intended order of physical priorities before the supervised surrogate is fitted \cite{Xue2025}.

Let \(\mathbf{x}_i\) denote the scan-order descriptor vector used by the surrogate model. This descriptor represents the completed track-order sequence through order-derived and geometry-aware sequence features used in the implementation, rather than through Abaqus field quantities. A supervised surrogate model \(f_{\text{sur}}\) was fitted to approximate the mapping from scan-order descriptors to the teacher-derived terminal reward:
\[
  \hat{r}_i = f_{\text{sur}}(\mathbf{x}_i; \phi)
\]
where \(\phi\) denotes the fitted model parameters. A histogram-based gradient boosting regressor, implemented using \texttt{HistGradientBoostingRegressor}, was used because the surrogate task is a nonlinear tabular regression problem over engineered scan-order descriptors \cite{Pedregosa2011}. After fitting, the model was frozen and used as the terminal reward function in the PPO environment:
\[
  R_{\text{surrogate}}(\sigma^{(N)}) = f_{\text{sur}}(\mathbf{x}(\sigma^{(N)}); \phi)
\]
where \(\sigma^{(N)}\) is a completed scan-order permutation and \(\mathbf{x}(\sigma^{(N)})\) is its descriptor representation.

The surrogate was fitted using the native 552-instance teacher-labelled pool and validated before PPO deployment. For the \(U_2\)--\(\mathrm{PEEQ}\)--\(\mathrm{Surface}_T\) hierarchical terminal-reward target, the validation statistics were Spearman correlation = 0.8786, Pearson correlation = 0.8863, MAE = 0.0942 and RMSE = 0.1379. Within-N Spearman correlations were 0.8164 for N12, 0.7940 for N16, 0.8449 for N24 and 0.9536 for N40. These statistics are reported here to define the surrogate environment used for PPO training; the physical performance of PPO-generated scan orders is assessed separately by Abaqus teacher validation.

During PPO training, the frozen surrogate returns a scalar reward only after a complete scan-order permutation has been generated. Intermediate rewards remain zero, consistent with the finite-horizon formulation in Section 2.2. The surrogate is also used to prioritise candidates within the PPO-generated rollout pool before Abaqus validation. It is not used as the final physical evaluator, and surrogate score alone is not treated as evidence of physical superiority. The surrogate defines the reinforcement-learning reward environment and candidate-prioritisation signal; Abaqus-extracted teacher metrics define the physical evidence.

\begin{figure}[tb]
  \centering
  \includegraphics[width=0.95\textwidth]{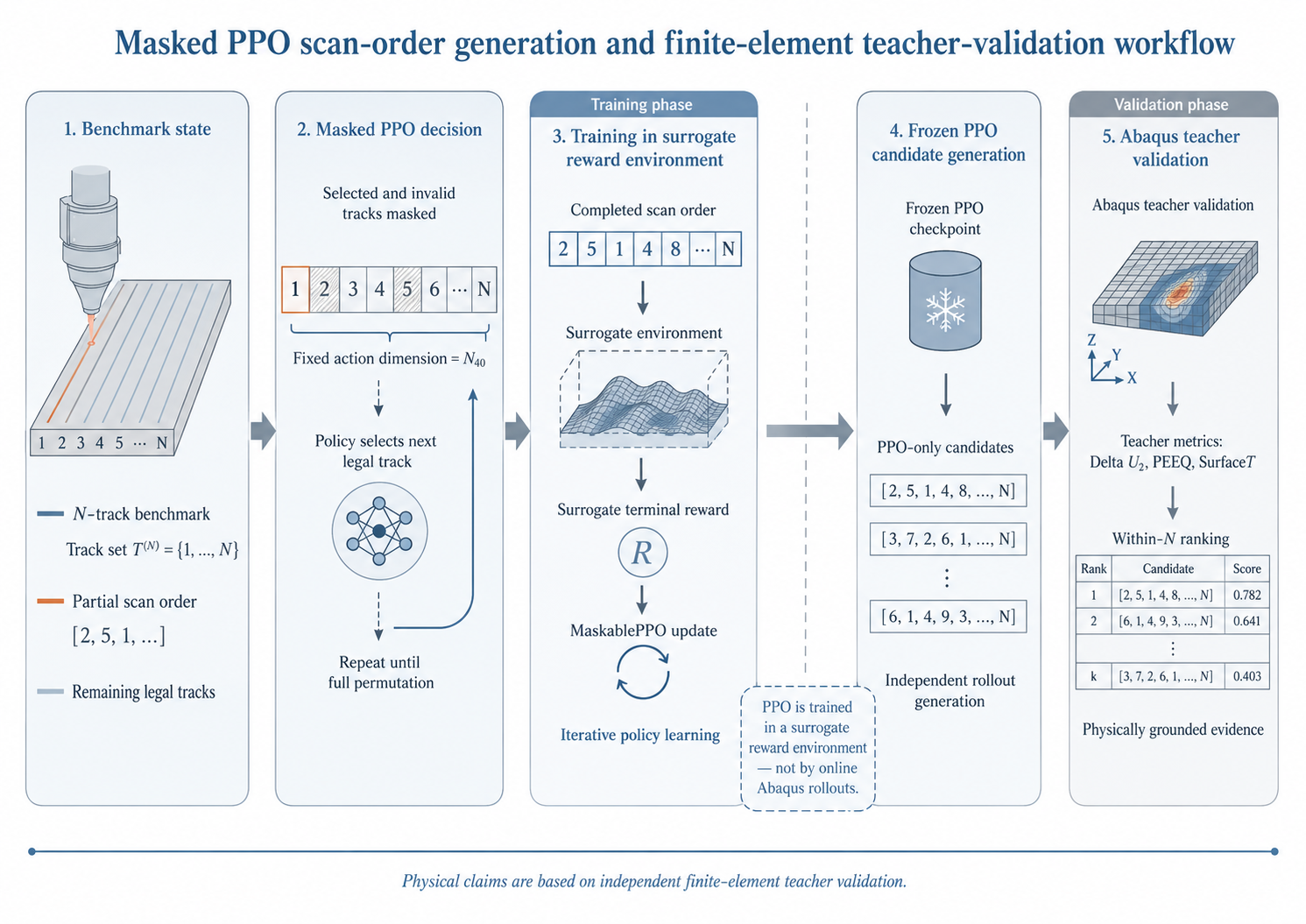}
  \caption{Bilevel surrogate-trained MaskablePPO scan-order generation and Abaqus teacher-validation workflow. Native Abaqus teacher-labelled scan-order data are transformed into a \(U_2\)--\(\mathrm{PEEQ}\)--\(\mathrm{Surface}_T\) hierarchical terminal-reward target. A histogram-based gradient boosting surrogate is fitted to this target to learn the teacher-labelled response landscape and is frozen as the PPO terminal-reward environment. MaskablePPO generates legal scan-order permutations under action masking. Selected PPO-only candidates are then converted into Abaqus cases, and their final physical ranking is determined from independently extracted teacher metrics rather than surrogate predictions.}
  \label{fig:2.2}
\end{figure}

\subsection{MaskablePPO policy training and PPO-only candidate generation}

MaskablePPO was used as the policy-gradient generator for scan-order permutations in the frozen surrogate terminal-reward environment defined in Section 2.3 \cite{Schulman2017,Raffin2021,Huang2022}. This stage corresponds to the policy-generation layer of the bilevel workflow: the surrogate-assisted teacher-guided optimisation loop provides the learned terminal-reward landscape, while MaskablePPO learns an executable sequential decision policy within that landscape. The PPO experiment is therefore designed to test whether a frozen policy checkpoint can autonomously generate legal and teacher-evaluable scan-order candidates, rather than to reproduce the full surrogate-assisted optimisation loop. This algorithmic choice is aligned with the structure of the scan-order problem: the action space is discrete, the episode is finite-horizon, and the admissible decision set changes after every selected track. Rather than penalising illegal actions after they occur, invalid-action masking removes infeasible sequence continuations before the policy distribution is sampled or optimised \cite{Huang2022}. This is essential for scan-order generation, because a valid output must be a complete permutation in which each deposition track appears exactly once.

The policy was implemented using an MlpPolicy and operated on the fixed N40-compatible action interface introduced in Section 2.2. For each N-track episode, the action mask invalidated two classes of actions: track indices outside the current benchmark and tracks already present in the partial scan order. Let \(z_{j,t}\) denote the unnormalised policy logit for selecting track \(j\) at decision step \(t\). The masked logit is defined as:
\[
  \tilde{z}_{j,t} =
  \begin{cases}
    z_{j,t} & \text{if } M_{j,t} = 1, \\
    -\infty & \text{if } M_{j,t} = 0.
  \end{cases}
\]
The resulting masked policy distribution is:
\[
  \pi_\theta(a_t = j \mid s_t, M_t) =
  \frac{\exp(\tilde{z}_{j,t})}
  {\sum_{\ell \in \mathcal{I}_{40}} \exp(\tilde{z}_{\ell,t})},
\]
where \(\mathcal{I}_{40}\) denotes the fixed N40-compatible action index set. The action mask is therefore part of the policy-optimisation mechanism rather than a post-processing filter, ensuring that probability mass is assigned only over admissible scan-order continuations \cite{Huang2022}.

The PPO update was applied to this masked policy distribution following the clipped policy-gradient objective introduced in proximal policy optimisation \cite{Schulman2017}. For a sampled transition, the probability ratio is:
\[
  \rho_t(\theta) = \frac{\pi_\theta(a_t \mid s_t, M_t)}{\pi_{\theta_{\text{old}}}(a_t \mid s_t, M_t)},
\]
and the clipped policy objective is:
\[
  L_{\text{PPO}}(\theta) = \mathbb{E}_t \Bigl[ \min\bigl( \rho_t(\theta) A_t,\ \text{clip}(\rho_t(\theta), 1-\varepsilon, 1+\varepsilon) A_t \bigr) \Bigr],
\]
where \(A_t\) denotes the estimated advantage and \(\varepsilon\) is the PPO clipping parameter. The return was sparse and terminal. Intermediate rewards were zero, and the completed scan-order permutation received the surrogate terminal reward:
\[
  R_t = 0 \quad \text{for } t < N, \qquad R_N = R_{\text{surrogate}}(\sigma^{(N)}).
\]
The corresponding policy objective can be written as:
\[
  \theta^* = \arg\max_\theta \mathbb{E}_{\sigma^{(N)} \sim \pi_\theta} \bigl[ R_{\text{surrogate}}(\sigma^{(N)}) \bigr]
\]
subject to \(\sigma^{(N)}\) being a valid permutation of the current N-track benchmark. This formulation makes the PPO policy a generator of legal scan-order candidates in a surrogate reward environment, rather than an online controller coupled directly to Abaqus.

The final MaskablePPO training run completed 200352 timesteps and contained 72,937 trainable parameters. After training, the policy checkpoint was frozen. Candidate generation and subsequent finite-element validation were performed without further policy updates.

PPO-only candidate generation was carried out by inference from the frozen checkpoint. Deterministic and stochastic rollout modes were used to form a policy-generated rollout pool for each native track-count setting. Let \(S_{\text{PPO}}^{(N)}\) denote the set of candidate scan orders generated by the frozen PPO policy for a given \(N\). Candidate prioritisation was then performed only within this PPO-generated set using the surrogate terminal reward. The selected subset for each \(N\) is denoted as \(C_{\text{PPO}}^{(N)}\), with:
\[
  C_{\text{PPO}}^{(N)} \subset S_{\text{PPO}}^{(N)}, \qquad |C_{\text{PPO}}^{(N)}| = 8.
\]
The complete PPO validation batch was formed by combining the four native-N subsets:
\[
  C_{\text{PPO}} = C_{\text{PPO}}^{(12)} \cup C_{\text{PPO}}^{(16)} \cup C_{\text{PPO}}^{(24)} \cup C_{\text{PPO}}^{(40)},
\]
with
\[
  |C_{\text{PPO}}^{(12)}| = |C_{\text{PPO}}^{(16)}| = |C_{\text{PPO}}^{(24)}| = |C_{\text{PPO}}^{(40)}| = 8
\]
and therefore \(|C_{\text{PPO}}| = 32\).

This selection rule was applied only within the PPO-generated rollout pool. No engineering baseline, predefined heuristic scan order, manual repair, local order mutation or post-hoc replacement was inserted into the PPO validation batch. Thus, the selected batch tests the output of the trained PPO policy-generation mechanism itself.

Novelty was assessed by comparing canonical scan-order identifiers against the native 552-instance teacher-labelled reference pool. Of the 32 selected PPO candidates, 31 were novel relative to this reference pool. One N12 candidate matched an existing teacher-labelled scan order and was retained as a recovery anchor rather than treated as a novel PPO discovery. Exact candidate identifiers are reported in the evidence tables rather than in the method definition.

The output of this stage is therefore not a surrogate-validated optimum and not the output of the full surrogate-assisted teacher-guided optimiser, but a frozen-checkpoint PPO candidate batch prepared for independent finite-element teacher validation. The surrogate defines the training and candidate-prioritisation environment; Abaqus-extracted teacher metrics define the physical evidence used for validation and ranking. This protocol evaluates whether the PPO policy has distilled useful scan-order structure from the learned reward landscape into executable candidate generation, not whether PPO surpasses the mature surrogate-assisted best records.

\subsection{Automated Abaqus teacher validation and label admission}

An automated Abaqus teacher-validation layer was used to convert PPO-generated scan orders into reproducible finite-element evidence. This layer provides the methodological separation between policy generation and physical assessment: PPO proposes candidate permutations, whereas Abaqus supplies the terminal thermo-mechanical teacher metrics used for final physical comparison, consistent with the use of finite-element thermo-mechanical workflows for DED process modelling \cite{Stender2018}. No PPO-generated candidate is admitted into the physical evidence set on the basis of surrogate score alone.

For a selected PPO scan order \(\sigma^{(N)}\), the validation layer defines a finite-element teacher mapping:
\[
  M_{\mathrm{FEA}} : \sigma^{(N)} \mapsto \mathbf{y}_{\mathrm{teacher}}^{(N)},
\]
where \(M_{\mathrm{FEA}}\) denotes the automated Abaqus validation procedure and \(\mathbf{y}_{\mathrm{teacher}}^{(N)}\) is the extracted terminal response vector. In implementation, this mapping includes CAE/INP case generation, Abaqus thermo-mechanical solution, ODB post-processing and terminal metric extraction, following the broader use of finite-element thermo-mechanical workflows for DED process modelling \cite{Stender2018}.

Label admission was controlled by numerical-completion and data-availability gates. A candidate was admitted into the validated set only when the finite-element case was generated successfully, the Abaqus analysis completed, the ODB file was available and readable, and the required terminal metrics were extracted. Failed, incomplete or non-readable cases were retained in the execution record but excluded from teacher-labelled performance comparison. This procedure ensures that the final validation set contains only auditable finite-element responses rather than assumed solver outcomes.

The admitted teacher response contains the same physical quantities used throughout the study: vertical displacement range, maximum equivalent plastic strain, surface tensile-stress index and diagnostic stress measures. These Abaqus-extracted metrics define the physical evidence used to rank PPO-generated scan orders against the native teacher-labelled reference pool.

\begin{landscape}
  \begin{figure}[p]
    \centering
    \includegraphics[width=\linewidth]{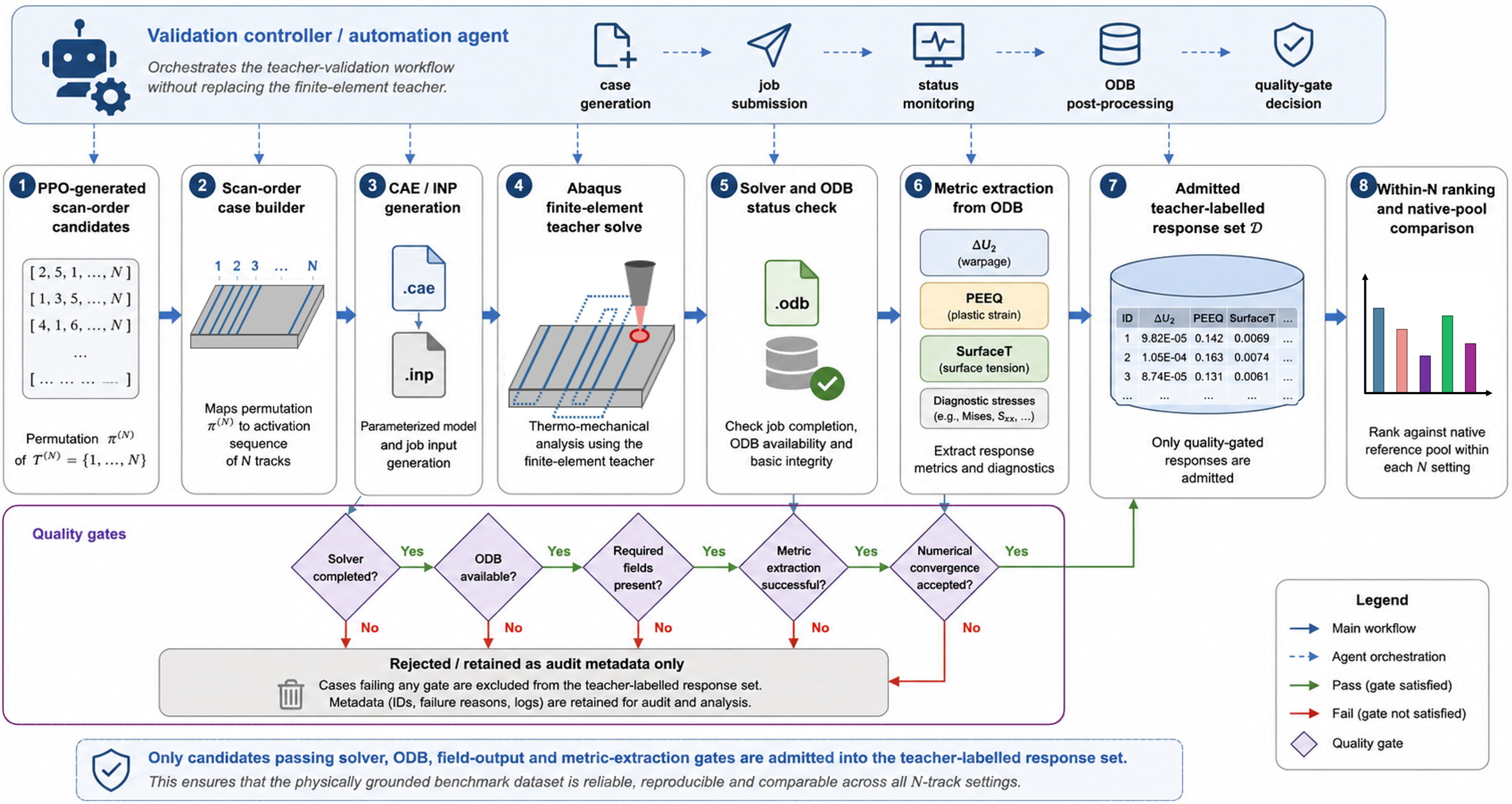}
    \caption{Automated Abaqus teacher-validation and label-admission workflow. PPO-generated scan-order candidates are converted into Abaqus CAE/INP cases, solved using the finite-element teacher, post-processed from ODB files and admitted into the validation set only after passing solver-completion, ODB-availability and metric-extraction gates. The admitted teacher-labelled responses, rather than surrogate predictions, provide the physical basis for ranking PPO-generated candidates against the native reference pool.}
    \label{fig:2.3}
  \end{figure}
\end{landscape}

\section{Results}







\subsection{Finite-element teacher-labelled response landscape across native and auxiliary track-count settings}

Before evaluating the PPO-generated scan orders, the finite-element response landscape used as the physical reference space must be established. The PPO candidates are not interpreted in isolation; they are assessed against a teacher-labelled distribution of warpage, plasticity and stress-related responses obtained under controlled LDED scan-order variation. This landscape supports the reward hierarchy, the surrogate terminal-reward model and the subsequent teacher-metric ranking of PPO-generated candidates.

The full response-landscape pool contains 884 Abaqus-labelled scan-order cases. The native subset contains 552 cases across the four primary track-count settings: 78 cases for N12, 78 cases for N16, 190 cases for N24 and 206 cases for N40. These native cases define the reference pool used later for PPO performance ranking. The remaining 332 fixed-N32 cases are retained as an auxiliary teacher-labelled response set. They are used to strengthen the response-landscape interpretation, especially for warpage-related displacement and plasticity response, but are excluded from the native PPO ranking boundary because the PPO validation batch contains only N12, N16, N24 and N40 candidates.

\begin{table}[htbp]
  \centering
  \caption{Finite-element teacher-labelled evidence pools used in the Results analysis. Native N12, N16, N24 and N40 cases contribute both to the response-landscape analysis and the PPO ranking reference pool. Auxiliary N32 cases are retained for response-landscape analysis but excluded from the native PPO ranking boundary.}
  \label{tab:evidence-pools}
  \begin{tabular}{@{}lcc>{\centering\arraybackslash}p{2.8cm}>{\centering\arraybackslash}p{2.3cm}>{\raggedright\arraybackslash}p{5.5cm}@{}}
    \toprule
    Pool      & N  & Cases & Response landscape & PPO ranking & Metric boundary \\
    \midrule
    Native    & 12 & 78    & Included           & Included    & Common native metric definitions \\
    Native    & 16 & 78    & Included           & Included    & Common native metric definitions \\
    Native    & 24 & 190   & Included           & Included    & Common native metric definitions \\
    Native    & 40 & 206   & Included           & Included    & Common native metric definitions \\
    \midrule
    Auxiliary & 32 & 332   & Included           & Excluded    & U2/PEEQ retained; stress proxies isolated \\
    \midrule
    \textbf{Total} & \textemdash & \textbf{884} & \textemdash & \textemdash & \textemdash \\
    \bottomrule
  \end{tabular}
\end{table}

Warpage-related displacement is the most scale-sensitive teacher metric in the native response landscape. The median vertical displacement range increases from approximately \(3.953 \times 10^{-5}\) for N12 to \(4.952 \times 10^{-5}\) for N16, \(8.096 \times 10^{-5}\) for N24 and \(1.498 \times 10^{-4}\) for N40. The N40 cases also exhibit a broader upper tail, indicating amplified sequence-dependent dispersion in the terminal warpage response as the scan-order decision horizon becomes longer. This scale-dependent dispersion is consistent with prior DED studies showing that distortion and residual-stress responses are sensitive to substrate geometry, thermal conditions and part geometry \cite{Corbin2018,Lu2019}. This behaviour motivates within-N comparison: direct global ranking across different track-count settings would conflate scan-order quality with differences caused by sequence length and benchmark geometry.

The equivalent plastic strain response is more stable across the native settings. The native median \(\mathrm{PEEQ}\) values are approximately 0.1485 for N12, 0.1520 for N16, 0.1573 for N24 and 0.1575 for N40. This narrower variation indicates that \(\mathrm{PEEQ}\) provides a consistent plasticity-related safety signal, but is less discriminative than warpage-related displacement as a primary ranking metric. The auxiliary N32 cases enlarge the warpage–plasticity evidence base and support the interpretation of \(\mathrm{PEEQ}\) as a safety filter rather than the main optimisation driver.

Stress-related responses are interpreted within a stricter comparability boundary. For the native N12, N16, N24 and N40 cases, \(\mathrm{Surface}_T\) and \(\mathrm{Mises}\) stress are available on a common post-processing basis. The median \(\mathrm{Surface}_T\) value increases moderately from approximately 581.3 MPa for N12 to approximately 583.3 MPa for N40, with N40 showing the largest high-stress tail. This residual-stress-related variation gives \(\mathrm{Surface}_T\) value as a secondary ranking signal after warpage and plasticity constraints have been considered. By contrast, \(\mathrm{Mises}\) stress remains tightly clustered near 580 MPa across the native cases, indicating limited rank-discriminative sensitivity within the native response table. It is therefore retained as a diagnostic stress response rather than used as a primary scan-order ranking signal.

The auxiliary N32 stress-related values are not merged into the native MPa-scale stress panels because their source-unit and extraction-definition compatibility with the native \(\mathrm{Surface}_T\) and \(\mathrm{Mises}\) definitions was not established. This separation allows the N32 dataset to strengthen the broader response-landscape analysis without weakening the numerical consistency of the native stress comparison.

\begin{figure}[tb]
  \centering
  \includegraphics[width=0.95\textwidth]{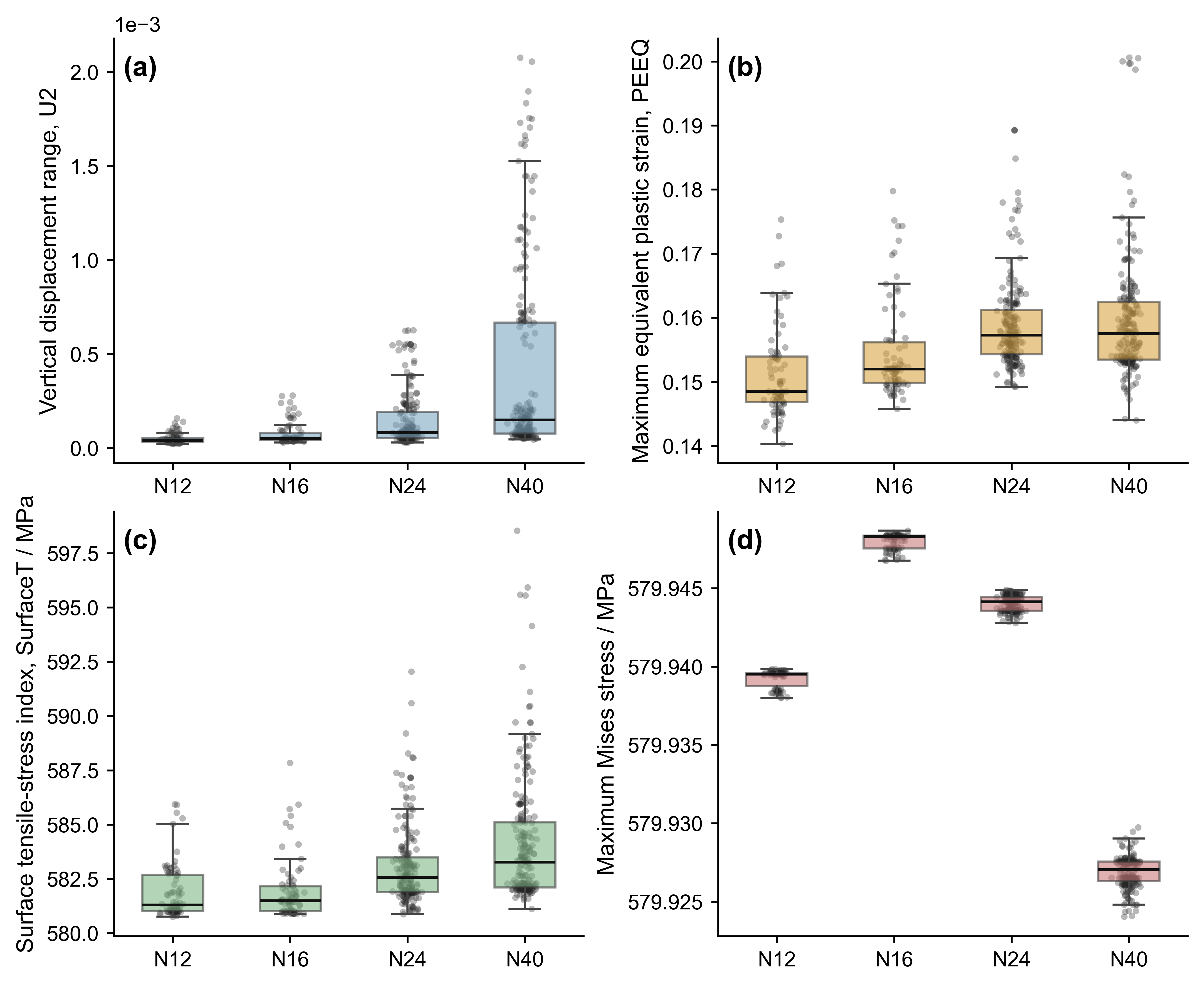}
  \caption{Teacher-labelled response distributions across native and auxiliary track-count settings. The figure shows \(U_2\) and \(\mathrm{PEEQ}\) distributions for N12, N16, N24, auxiliary N32 and N40, and native MPa-scale \(\mathrm{Surface}_T\) and \(\mathrm{Mises}\) distributions for N12, N16, N24 and N40. The auxiliary N32 set is retained for response-landscape interpretation, while stress-related N32 quantities are shown separately or excluded from native MPa-scale stress panels where extraction semantics are not directly compatible.}
  \label{fig:3.X}
\end{figure}

This response landscape defines the physical coordinate system for the subsequent PPO analysis. The 884-case pool supports the empirical interpretation of warpage, plasticity and stress-related behaviour, whereas the native 552-case subset provides the clean reference boundary for ranking PPO-generated scan orders. It also provides the empirical basis for deriving the \(U_2\)-first, \(\mathrm{PEEQ}\)-safe and \(\mathrm{Surface}_T\)-secondary reward hierarchy analysed next.

\subsection{Empirical reward hierarchy derived from teacher-labelled response statistics}

The surrogate reward environment used for PPO training requires a terminal reward that preserves the physical priority of the teacher-labelled responses. The response landscape shows that warpage-related displacement, plastic-strain accumulation and stress-related response are not interchangeable objectives. A flat weighted scalarisation would introduce compensatory behaviour, allowing improvement in a secondary stress-related response to offset deterioration in a primary geometric response \cite{Skalse2022}. This is undesirable for scan-order decision-making, because a candidate with excessive warpage should not be promoted simply because it has a favourable residual-stress proxy. The reward hierarchy is therefore derived from the teacher-labelled response statistics rather than assigned as an arbitrary weighted sum.

The primary layer is defined by \(U_2\), which represents the vertical displacement range and serves as the indicator of warpage-related geometric distortion. Section~3.1 showed that \(U_2\) is the most scale-sensitive response, with both its median value and upper-tail dispersion increasing with track count. This makes \(U_2\) the most discriminative metric for geometric admissibility. Consequently, a favourable \(\mathrm{Surface}_T\) value is not allowed to compensate for an unfavourable \(U_2\) response.

The \(U_2\)--\(\mathrm{Surface}_T\) relationship supports this ordering. Across the native teacher-labelled cases, the raw Spearman correlation between \(U_2\) and \(\mathrm{Surface}_T\) is 0.5414, but decreases to 0.3963 after within-N normalisation. The correlation is also strongly track-count dependent, with values of 0.1690 for N12, 0.4686 for N16, 0.2136 for N24 and 0.6577 for N40. These statistics show that \(U_2\) and \(\mathrm{Surface}_T\) are related, but not equivalent, and that their relationship cannot be represented by a single global scalar trend.

The response-mismatch cases expose the failure mode of a \(\mathrm{Surface}_T\)-first rule. In the within-N normalised multi-response space, 15 native candidates show low \(\mathrm{Surface}_T\) but high \(U_2\), while 6 candidates show low \(U_2\) but high \(\mathrm{Surface}_T\). The former group could be incorrectly favoured by an unconstrained stress-first criterion despite remaining unfavourable in the primary warpage response. The latter group shows the converse: low-warpage candidates can still require stress-based secondary ranking. This mismatch motivates a gated reward structure in which \(U_2\) defines the primary admissibility layer and \(\mathrm{Surface}_T\) is activated only after the geometric response is acceptable or near-admissible \cite{Tercan2024}.

PEEQ forms the second layer of the hierarchy. In the native teacher-labelled set, 532 of 552 cases satisfy the analysis-defined \(\mathrm{PEEQ}\)-safe condition, indicating that \(\mathrm{PEEQ}\) is not the most restrictive response in the current landscape. It remains necessary, however, as a plasticity-safety filter: a low-\(U_2\) scan order should not be promoted if it produces comparatively unfavourable equivalent plastic strain. \(\mathrm{PEEQ}\) therefore constrains the admissible response region before secondary stress-related ranking is applied.

\(\mathrm{Surface}_T\) retains its role after the \(U_2\) and \(\mathrm{PEEQ}\) gates. The combined-safe region contains 405 native candidates, and the near-admissible region contains 478 candidates. Within the combined-safe region, \(\mathrm{Surface}_T\) still spans an all-native range of approximately 17.7 MPa. This residual variation shows that stress-related optimisation remains meaningful inside the admissible response region. \(\mathrm{Surface}_T\) is therefore not discarded; it provides conditional secondary ranking after warpage and plasticity requirements have been satisfied.

\begin{figure}[tb]
  \centering
  \includegraphics[width=0.95\textwidth]{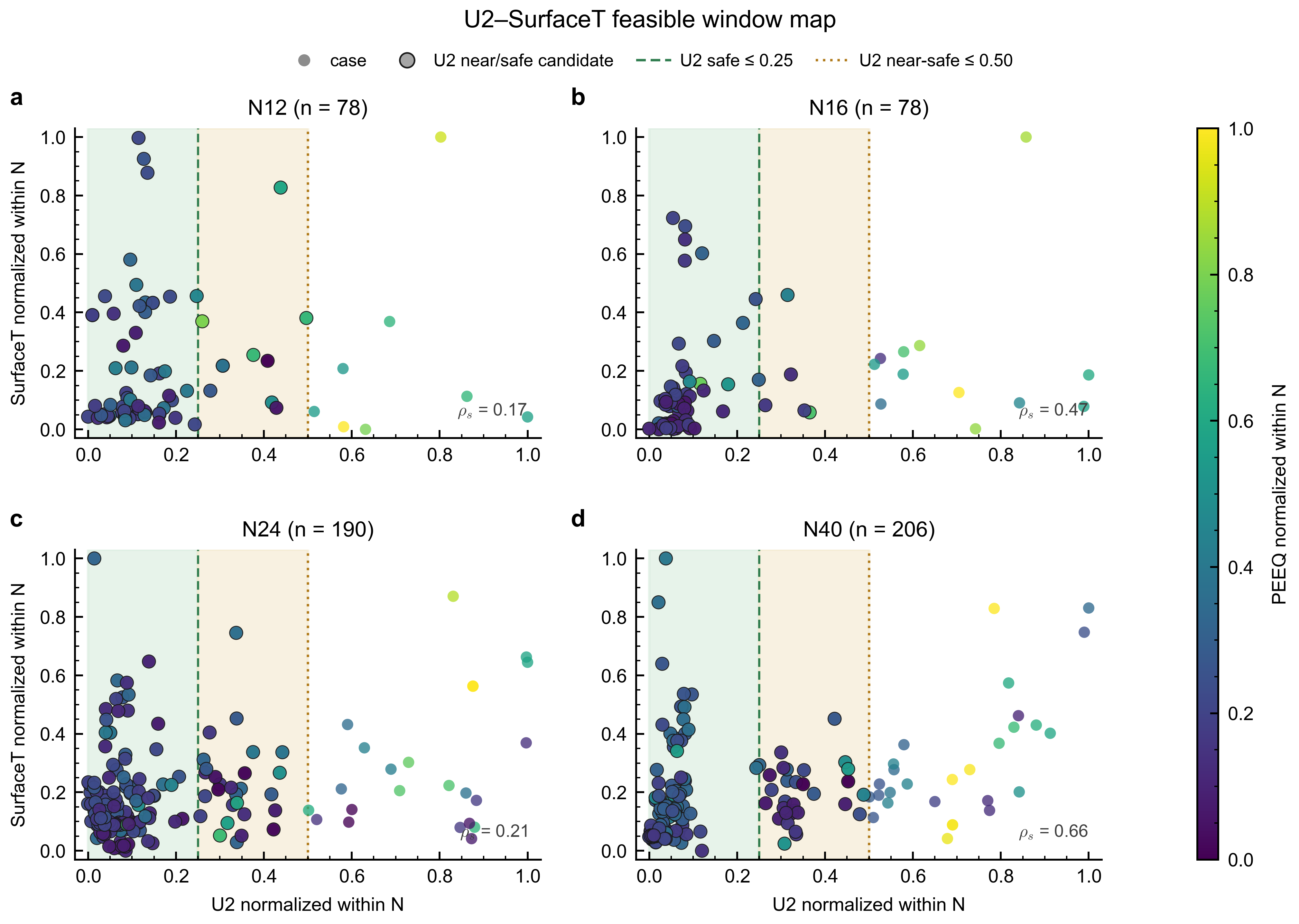}
  \caption{Native multi-N feasible-window map linking warpage admissibility and surface tensile-stress response.}
  \label{fig:3.1}
\end{figure}

\(\mathrm{Mises}\) stress is retained only diagnostically. Although available in the native teacher-labelled response table, it is tightly clustered across the native settings and has limited rank-discriminative sensitivity. Allowing such a weakly discriminative diagnostic response to drive the reward would obscure the more informative \(U_2\) and \(\mathrm{Surface}_T\) structure. \(\mathrm{Mises}\) is therefore reported for interpretation but excluded as a primary reward driver.

\begin{figure}[tb]
  \centering
  \includegraphics[width=0.95\textwidth]{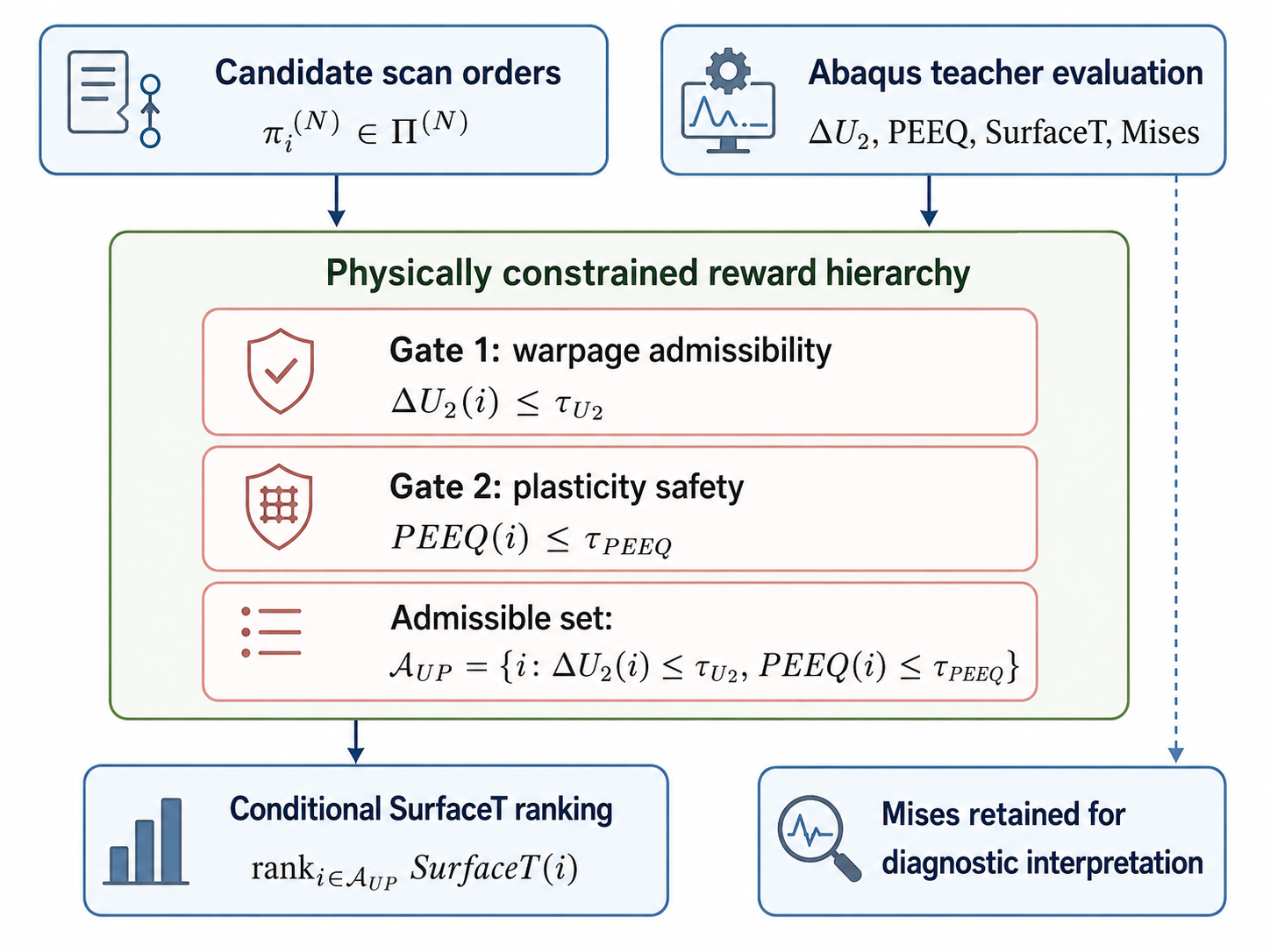}
  \caption{Physically constrained reward hierarchy derived from teacher-labelled response statistics. \(U_2\) defines the primary warpage-admissibility layer, \(\mathrm{PEEQ}\) acts as the plasticity-safety filter, and \(\mathrm{Surface}_T\) provides conditional secondary ranking inside the admissible or near-admissible region. \(\mathrm{Mises}\) stress is retained for diagnostic interpretation.}
  \label{fig:3.3}
\end{figure}

\begin{table}[tb]
  \centering
  \small
  \caption{Roles of Abaqus-derived teacher metrics in the reward hierarchy}
  \label{tab:3.2}
  \begin{tabularx}{0.95\textwidth}{@{}p{0.22\textwidth}XX@{}}
    \toprule
    \textbf{Metric} & \textbf{Role} & \textbf{Treatment} \\
    \midrule
    \(U_2\) & Warpage admissibility & Always active \\
    \(\mathrm{PEEQ}\) & Plasticity safety & Gate before secondary ranking \\
    \(\mathrm{Surface}_T\) & Residual-stress ranking & Active only after admissibility gates \\
    Mises & Stress diagnostic & Reported, not reward-driving \\
    \bottomrule
  \end{tabularx}
\end{table}

The teacher-labelled response statistics therefore support a gated hierarchy rather than a flat scalarisation: \(U_2\) establishes geometric admissibility, \(\mathrm{PEEQ}\) enforces plasticity safety, and \(\mathrm{Surface}_T\) ranks candidates conditionally within the admissible response region. This hierarchy defines the physical terminal-reward target used in the surrogate PPO training environment analysed next.

\subsection{Surrogate terminal-reward model validation}

The PPO policy was trained without online Abaqus interaction, so the surrogate terminal-reward model is a critical link between the finite-element teacher-labelled response landscape and policy-gradient learning. Its purpose is not to replace Abaqus, but to provide a computationally tractable reward environment that preserves enough teacher-derived rank structure for PPO training and candidate prioritisation \cite{Kim2026,Francon2020}.

The surrogate was fitted using the native teacher-labelled pool only. This pool contains the 552 native N12, N16, N24 and N40 cases described in Section 3.1. The auxiliary fixed-N32 cases support response-landscape interpretation, especially for warpage and plasticity trends, but are not part of the primary native surrogate-training boundary. This separation keeps the surrogate aligned with the native track-count settings used later for PPO candidate generation and teacher-metric ranking.

The supervised target was the physically gated lexicographic terminal reward derived from the teacher response statistics in Section 3.2. This reward target encodes the physical priority of the response hierarchy: \(U_2\) defines geometric admissibility, \(\mathrm{PEEQ}\) enforces plasticity safety, and \(\mathrm{Surface}_T\) provides secondary ranking inside the admissible or near-admissible response region. The surrogate therefore learns a terminal reward derived from finite-element responses, rather than predicting an intermediate thermal or mechanical field quantity.

A histogram-based gradient boosting surrogate was used to map scan-order descriptors to this teacher-derived terminal reward \cite{Pedregosa2011}. Across the native validation data, the model achieved a Spearman correlation of 0.8786 and a Pearson correlation of 0.8863, with MAE = 0.0942 and RMSE = 0.1379. These values indicate that the surrogate captures both the ranking structure and the magnitude trend of the teacher-derived reward sufficiently to support policy-gradient training.

\begin{figure}[htbp]
  \centering
  \includegraphics[width=0.95\textwidth]{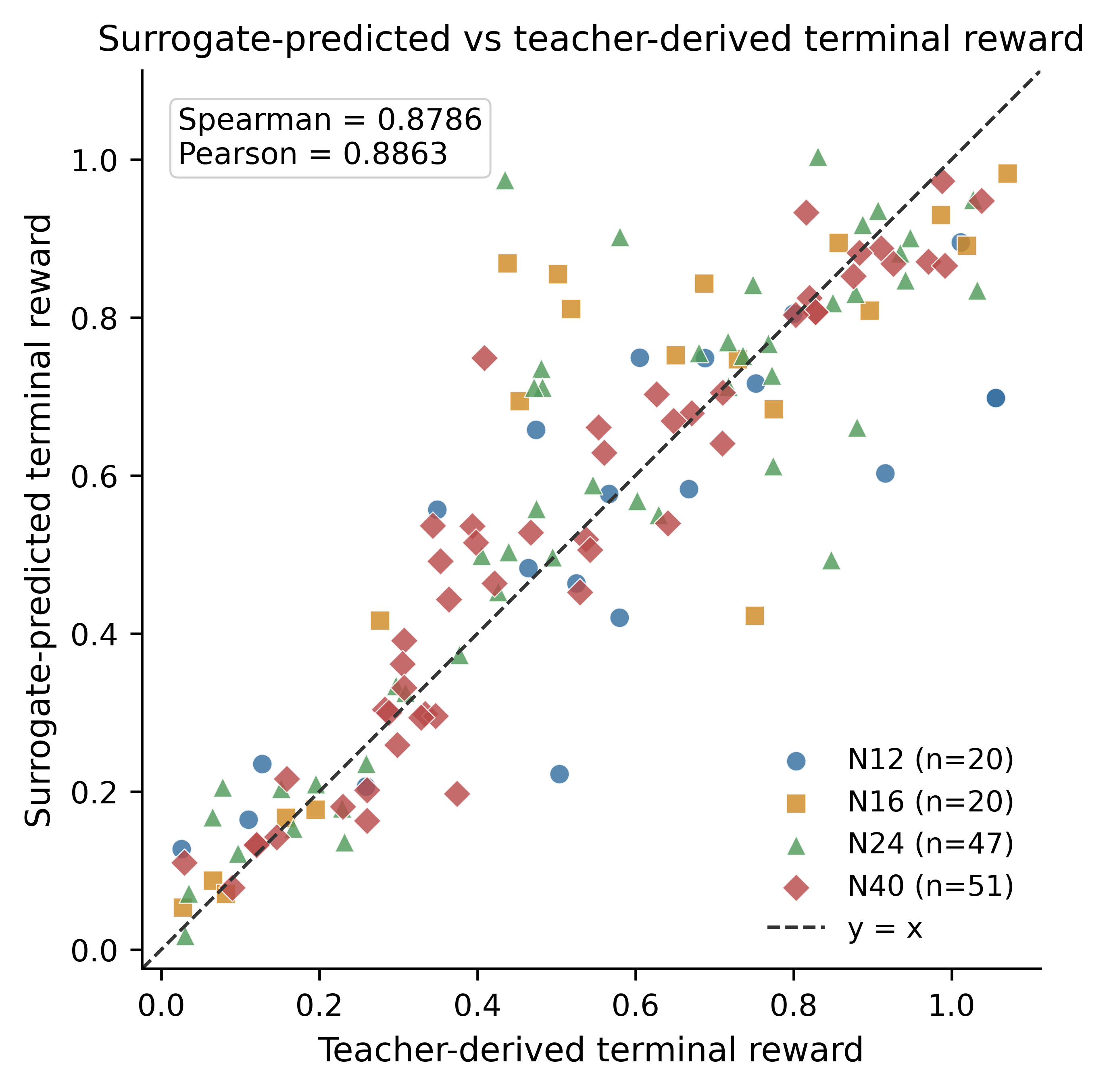}
  \caption{Surrogate-predicted terminal reward versus teacher-derived terminal reward. The figure should compare predicted and teacher-derived reward values for the native N12, N16, N24 and N40 validation data. The reported overall agreement is Spearman correlation = 0.8786 and Pearson correlation = 0.8863. The figure validates the surrogate as a PPO training environment, not as a substitute for final Abaqus teacher validation.}
  \label{fig:3.4}
\end{figure}

The within-\(N\) validation results further support this role. Spearman correlations remained positive across all native track-count settings: 0.8164 for N12, 0.7940 for N16, 0.8449 for N24 and 0.9536 for N40. The strongest passive validation agreement was obtained for N40, the largest native setting. This indicates that the descriptor-based surrogate retained useful ranking information within the teacher-labelled validation data even as the scan-order decision horizon increased. However, this validation result should not be interpreted as proof that PPO-generated N40 candidates will necessarily remain top-ranked after policy-induced sampling and Abaqus teacher validation. It supports the use of a common surrogate reward environment across N12, N16, N24 and N40, while leaving final PPO competitiveness to be determined by teacher-extracted metrics.

\begin{table}[htbp]
  \centering
  \caption{Validation metrics (Spearman, Pearson, MAE and RMSE) for the native pooled model (N12, N16, N24 and N40) and individual per-N models.}
  \label{tab:validation-metrics}
  \begin{tabular}{@{}>{\raggedright\arraybackslash}p{5.8cm}rrrr@{}}
    \toprule
    Validation scope                  & Spearman & Pearson & MAE    & RMSE   \\
    \midrule
    Native pooled N12, N16, N24 and N40     & 0.8786   & 0.8863  & 0.0942 & 0.1379 \\
    \midrule
    N12                               & 0.8164   & \textemdash & \textemdash & \textemdash \\
    N16                               & 0.7940   & \textemdash & \textemdash & \textemdash \\
    N24                               & 0.8449   & \textemdash & \textemdash & \textemdash \\
    N40                               & 0.9536   & \textemdash & \textemdash & \textemdash \\
    \bottomrule
  \end{tabular}
\end{table}

These results establish the surrogate as a rank-informative terminal-reward environment for MaskablePPO training. The surrogate provides efficient reward feedback during policy optimisation, but it does not define the final physical performance of PPO-generated scan orders \cite{Kim2026,Francon2020}. The following subsections therefore evaluate PPO-generated candidates using Abaqus-extracted teacher metrics rather than surrogate predictions.

\subsection{Abaqus teacher validation of PPO-generated scan orders}

The frozen MaskablePPO policy produced a PPO-only validation cohort of 32 scan-order candidates, with 8 candidates each for N12, N16, N24 and N40. No engineering baseline insertion, heuristic insertion, manual repair, local mutation or post-hoc replacement was applied. The cohort therefore provides a direct test of whether the frozen policy can generate scan-order permutations that are admissible to independent finite-element teacher validation, before any claim about ranking competitiveness is made.

All 32 PPO-generated scan orders passed the complete Abaqus teacher-validation chain. Each candidate was converted into an Abaqus CAE model, exported as an INP file, solved successfully, admitted to ODB-based post-processing and assigned terminal teacher metrics. The extracted fields included displacement \(U\), equivalent plastic strain \(\mathrm{PEEQ}\), stress \(S\) and temperature output through the available NT or NT11-equivalent record. The final cooling step was retained consistently across the validation cohort, with timePeriod = 1200.0, initialInc = 0.01 and maxInc = 60.0, so that terminal metrics were extracted from a common post-deposition response state.

This result establishes complete finite-element evaluability of the PPO-generated validation cohort. The candidates were not only legal discrete permutations under the policy mask, but also executable thermo-mechanical activation schedules under the independent Abaqus teacher workflow. This validation completion does not by itself imply performance superiority; rather, it provides the required teacher-labelled evidence base for ranking PPO-generated candidates against the native reference pool in the next subsection.

\subsection{PPO competitiveness against the native reference pool}

The teacher-validated PPO cohort was ranked against the native N12, N16, N24 and N40 reference pool only. The ranking boundary was restricted to the 552 native teacher-labelled cases described in Section 3.1; the auxiliary fixed-N32 cases were excluded because they do not belong to the PPO validation settings. This comparison evaluates PPO-generated competitiveness within the native finite-element reference landscape used to define the surrogate-training and validation boundary. It should be interpreted as a policy-generation test, not as a claim that PPO outperforms the full surrogate-assisted teacher-guided optimisation loop.

The ranking analysis was designed as a limited-rollout policy test rather than an exhaustive optimisation claim. For each native track-count setting, only 8 PPO-generated candidates were admitted to Abaqus teacher validation. No engineering baseline insertion, heuristic insertion, manual repair, local mutation or post-hoc replacement was applied. The resulting ranks therefore measure whether the frozen policy could concentrate a small number of independently teacher-validated candidates into competitive regions of the native response landscape. The strongest records in this reference landscape remain those obtained by the broader teacher-guided surrogate-assisted optimisation process.

The strongest rank concentration occurred in the smaller native settings. The best N12 candidate reached rank 6 within the N12 native reference pool, and the best N16 candidate reached rank 2 within the N16 native reference pool. Across the full 32-candidate PPO batch, 12 candidates satisfied the predefined top-k competitiveness criterion: 5 for N12, 4 for N16, 3 for N24 and 0 for N40. These results show that the policy retained meaningful competitiveness in the smaller-N regimes, particularly N12 and N16, under a limited teacher-validation budget.

\begin{table}[htbp]
  \centering
  \small
  \caption{PPO candidate statistics and native-reference record status across track-count settings. Top-\(k\) cases are counted using the predefined competitiveness views, whereas best rank reports the best within-\(N\) lexicographic reference rank reached by the PPO batch.}
  \label{tab:ppo-candidates}
  \begin{tabularx}{\textwidth}{@{}lccccX@{}}
    \toprule
    \textbf{Setting} & \textbf{Reference cases} & \textbf{PPO cases} & \textbf{Top-\(k\) cases} & \textbf{Best lex. rank} & \textbf{New record} \\
    \midrule
    N12 & 78  & 8 & 5 & 6   & No \\
    N16 & 78  & 8 & 4 & 2   & No \\
    N24 & 190 & 8 & 3 & 134 & No \\
    N40 & 206 & 8 & 0 & 147 & No \\
    \midrule
    Total & 552 & 32 & 12 & \textemdash & No \\
    \bottomrule
  \end{tabularx}
\end{table}

The rank concentration weakened as the scan-order horizon increased. For N24, the best PPO candidate reached lexicographic rank 134 within the N24 native reference pool, although 3 PPO candidates still met the top-\(k\) competitiveness criterion under the predefined metric/objective evaluation views. For N40, the best PPO candidate reached lexicographic rank 147, and no PPO-generated candidate entered the top-\(k\) competitive region. This \(N\)-dependent degradation is consistent with the increasing difficulty of learning rank-informative generation rules from sparse teacher-labelled supervision as the factorial scan-order space expands. It defines a high-\(N\) limitation of the present frozen policy, rather than a failure of finite-element evaluability, because all N40 candidates were successfully admitted to Abaqus teacher validation.

No PPO-generated candidate established a new native-reference record. This result sets the claim boundary clearly: the frozen policy is not presented as globally optimal, all-\(N\) dominant, capable of solving arbitrary-\(N\) scan-order optimisation or superior to the mature surrogate-assisted teacher-guided optimiser. Its demonstrated value is narrower but still meaningful. Under a PPO-only generation protocol and independent Abaqus teacher validation, the policy generated top-\(k\) competitive scan orders for N12 and N16, retained limited competitiveness for N24, and exposed a high-\(N\) reliability boundary at N40.

The PPO batch therefore supports a bounded competitiveness claim. It did not improve the existing native-reference records, but it produced teacher-validated competitive candidates in the smaller native regimes without post-hoc engineering repair. This result supports PPO as a teacher-validated policy-generation layer that can distil useful scan-order structure from the learned surrogate reward landscape, rather than as a replacement for the strongest surrogate-assisted optimisation loop. This \(N\)-dependent performance profile motivates the surrogate-to-teacher alignment analysis in the next subsection, where the reliability boundary of surrogate-guided PPO candidate selection is examined directly.

\subsection{Surrogate-to-teacher alignment and claim boundary}

The PPO-generated validation cohort provides a stricter test of surrogate reliability than the held-out validation reported in Section 3.3. The earlier validation assessed the surrogate on native teacher-labelled cases drawn from the reference pool. By contrast, the PPO cohort is policy-induced: the frozen policy selects candidates from regions favoured by the surrogate reward landscape. The evaluation therefore shifts from passive validation to surrogate assessment under optimisation pressure, where local prediction errors can be amplified by policy selection.

For the 32 PPO-generated candidates, the alignment between surrogate-predicted reward and Abaqus-derived teacher performance was weak but positive. The surrogate-to-teacher comparison gave Spearman correlation = 0.2790 and Pearson correlation = 0.2092. Within the predefined elite-prediction window, two PPO candidates were confirmed as high-performing by the Abaqus teacher metrics, whereas one candidate was overestimated by the surrogate. These results show that the surrogate retained some candidate-prioritisation signal, but its reliability was substantially weaker for policy-generated candidates than for passive validation on the native reference pool.

\begin{figure}[htbp]
  \centering
  \includegraphics[width=0.95\textwidth]{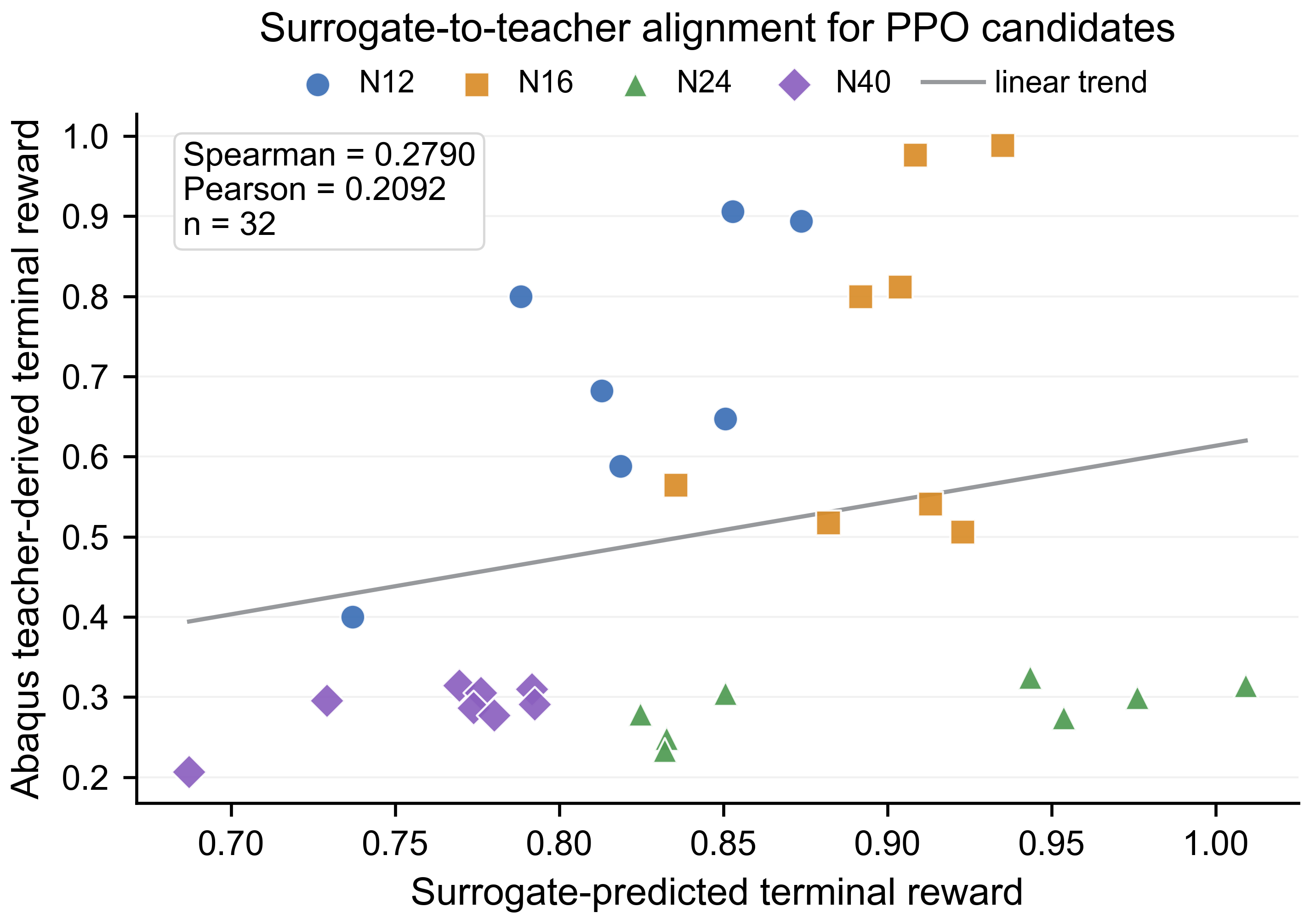}
  \caption{Surrogate-to-teacher alignment for PPO-generated candidates. The figure compares surrogate-predicted terminal reward with Abaqus-derived teacher performance for the 32 teacher-validated PPO candidates. The weak but positive alignment, with Spearman correlation = 0.2790 and Pearson correlation = 0.2092, indicates that the surrogate provides a useful candidate-prioritisation signal under policy-induced sampling but cannot serve as the final physical evaluator.}
  \label{fig:3.8}
\end{figure}

This alignment drop helps explain the performance pattern observed in Section 3.5. The surrogate was sufficiently informative to support PPO training and to generate competitive candidates in the smaller native settings. However, when the policy exploited the surrogate landscape, teacher-confirmed ranking was not preserved strongly enough to produce new native-reference records or all-\(N\) dominance. The issue is not finite-element infeasibility, because all PPO candidates were successfully admitted to Abaqus teacher validation. Rather, the limitation lies in the imperfect transfer from surrogate preference to teacher-confirmed performance under policy-induced selection.

The overestimated candidate exposes the failure mode that the teacher-validation layer is designed to intercept. A scan order can appear favourable under the surrogate terminal reward but fail to retain the same standing after Abaqus teacher extraction. This result directly supports the bilevel evidence structure used in this study: the surrogate can guide policy learning and candidate prioritisation, but physical-performance claims must remain anchored to Abaqus-extracted teacher metrics rather than surrogate scores.

\begin{table}[tb]
  \centering
  \small
  \caption{Claim boundary of surrogate-trained PPO scan-order generation.}
  \label{tab:3.6}
  \begin{tabularx}{\textwidth}{@{}>{\raggedright\arraybackslash}p{0.26\textwidth}>{\raggedright\arraybackslash}X>{\raggedright\arraybackslash}X@{}}
    \toprule
    \textbf{Claim level} & \textbf{Supported} & \textbf{Not claimed} \\
    \midrule
    Surrogate reward model & Rank-informative PPO training and candidate-prioritisation signal & Replacement for Abaqus evaluation \\
    PPO candidate validation & 32/32 PPO-only candidates admitted by Abaqus & Surrogate-only physical assessment \\
    PPO optimisation outcome & Bounded small-\(N\)/top-\(k\) competitiveness & New records, all-\(N\) dominance or arbitrary-\(N\) optimisation \\
    PPO relative to surrogate-assisted optimisation & Policy-generation layer distilled from the learned reward landscape & Superiority over the mature surrogate-assisted optimiser \\
    \bottomrule
  \end{tabularx}
\end{table}

The combined evidence defines a bounded but defensible claim. The PPO batch did not establish new native-reference records and did not demonstrate all-\(N\) dominance. It did, however, generate teacher-validated competitive candidates in the smaller native settings without heuristic insertion, manual repair or post-hoc replacement. Together, these results support teacher-validated PPO policy generation with bounded small-\(N\)/top-\(k\) competitiveness, not record-level optimisation, arbitrary-\(N\) scan-order solution or superiority over the mature surrogate-assisted optimisation loop.

\section{Discussion}






\subsection{Finite-element-grounded PPO scan-order generation}

The main methodological implication of the present results is that LDED scan-order design can be treated as finite-element-grounded policy generation, rather than only as a comparison among a small set of predefined scan strategies. Conventional scan-strategy studies usually begin with human-designed rules, such as raster, alternating, centre-out or region-based scanning, and then compare their thermo-mechanical consequences \cite{Dar2025,Ding2024}. The present framework extends this logic by adding a learned policy-generation route: scan orders are generated by a frozen MaskablePPO policy under the permutation constraint and are then admitted to finite-element teacher validation. The policy therefore synthesises candidate spatio-temporal activation sequences instead of selecting only from a fixed catalogue of scan rules.

This formulation is physically motivated by the delayed and non-local nature of scan-order effects. The deposition of one track modifies the thermal state inherited by subsequent tracks, alters the return time to neighbouring regions and reshapes the cumulative residual-strain, distortion and stress-relaxation fields after the full sequence is completed \cite{Caiazzo2022,Denlinger2015}. The quality of a local action cannot be judged reliably from local geometry alone, because its consequence depends on the remainder of the deposition history. This is why the reward must be terminal and finite-element grounded: the relevant manufacturing response is produced by the completed thermo-mechanical history of the permutation.

The value of the PPO layer is therefore not simply that it can propose scan orders, but that those proposals can be subjected to the same Abaqus teacher evaluation as conventional strategies and surrogate-assisted candidates. This changes the role of reinforcement learning in scan-order design. Instead of using PPO as a surrogate-only optimiser, the policy is used to expand the set of candidate orderings that can be physically interrogated. The bounded competitiveness observed in the native results should be interpreted in this light: the policy did not solve the full combinatorial optimisation problem and did not exceed the mature surrogate-assisted best records, but it generated Abaqus-validated candidates that entered competitive regions of the native response landscape under a limited rollout budget.

An auxiliary fixed-N32 comparison illustrates the same distinction from a legacy benchmark perspective. In that benchmark, learned scan-order candidates were compared with several conventional scan strategies using warpage and \(\mathrm{S}_{11}\) stress responses. This comparison is not part of the native PPO ranking analysis in Section~3.5, because the primary PPO validation boundary is restricted to N12, N16, N24 and N40. It is retained only as auxiliary evidence that learned scan-order generation can produce response profiles distinct from predefined scan rules under a fixed-\(N\) setting.

\begin{landscape}
  \begin{figure}[p]
    \centering
    \includegraphics[width=0.90\linewidth,height=0.78\textheight,keepaspectratio]{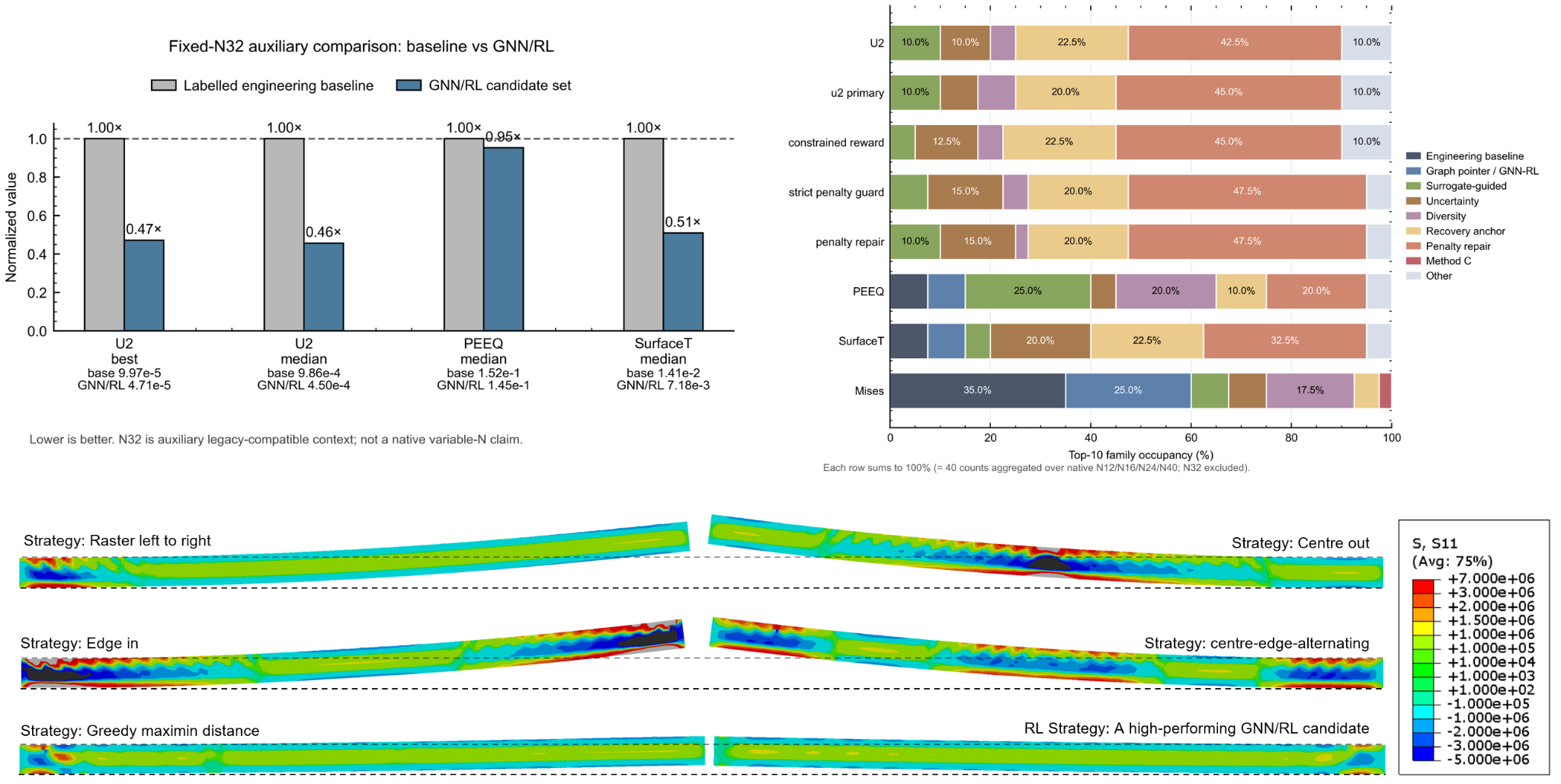}
    \caption{Auxiliary and diagnostic evidence for learned scan-order generation across fixed-N32 and native multi-\(N\) settings. (a) Fixed-N32 auxiliary comparison between the labelled engineering baseline and learned candidate set, showing normalised \(U_2\), \(\mathrm{PEEQ}\) and \(\mathrm{Surface}_T\) responses; lower values indicate better physical response. (b) Top-10 family occupancy aggregated over the native N12, N16, N24 and N40 teacher-labelled response sets, showing which strategy families populate the best-performing regions under different objective views; N32 is excluded from this native multi-\(N\) summary. (c) Representative fixed-N32 Abaqus teacher responses for conventional scan strategies and a learned candidate, visualised using deformed warpage profiles and \(\mathrm{S}_{11}\) stress contours. The fixed-N32 results are retained as auxiliary legacy-compatible evidence for distinguishing learned scan-order generation from predefined scan-strategy comparison, whereas native PPO ranking claims are based only on the N12, N16, N24 and N40 teacher-validated response sets reported in Section~3.5.}
    \label{fig:4.1}
  \end{figure}
\end{landscape}

This perspective sets up the remaining Discussion. Section 4.2 considers why the competitiveness of the policy-generated candidates is N-dependent. Section 4.3 then interprets the selected scan orders as physically meaningful sequence motifs rather than opaque permutations. The broader implication is that RL-for-AM can be used not only to tune continuous process variables, but also to generate discrete, physically testable toolpath orderings in a high-dimensional manufacturing design space.

\subsection{Bounded competitiveness and N-dependent reliability}

The competitiveness results should be interpreted as a limited-rollout policy-generation outcome rather than as evidence of exhaustive optimisation. The frozen PPO policy did not establish new native-reference records, nor did it exceed the mature surrogate-assisted best records. However, it generated teacher-validated candidates that entered competitive regions of the native response landscape in the smaller track-count settings. In an expensive finite-element setting, this distinction is important: the relevant question is not only whether a policy finds a new best case, but whether it can concentrate a small validation budget into physically admissible and competitive scan-order candidates.

The small-\(N\) results indicate that the policy captured useful ordering signals within the tested native boundary. With only 8 Abaqus-validated PPO candidates per track-count setting, the policy produced high-ranking candidates in N12 and N16 and multiple candidates satisfying the predefined competitiveness criterion. These ranks are not global ranks in the full permutation space. They instead show that the learned policy can act as a validation-budget-aware proposal mechanism within a finite teacher-labelled landscape, reducing dependence on blind enumeration or manual construction \cite{Bengio2021,Mazyavkina2021}.

The decline at larger \(N\) reflects a coupled data and search-space limitation. The nominal permutation space expands from \(16! \approx 2.1 \times 10^{13}\) to \(40! \approx 8.2 \times 10^{47}\), an increase of more than 34 orders of magnitude, while the native teacher-labelled reference data remain at only tens to hundreds of cases per \(N\). Under this imbalance, the surrogate-trained policy must infer ranking structure from an increasingly sparse physical evidence base, consistent with the broader difficulty of applying learning-based methods to large combinatorial optimisation spaces \cite{Bengio2021,Mazyavkina2021}. The observed shift from small-\(N\) competitiveness to no top-\(k\) competitiveness at N40 is therefore consistent with weaker surrogate-to-teacher transfer under policy-induced selection as the action horizon expands.

This interpretation separates executability from ranking competitiveness. The N40 candidates were not rejected by the finite-element workflow; they were successfully converted, solved and teacher-labelled. Their limitation was performance ranking, not finite-element executability. The mask-constrained policy and validation pipeline preserved scan-order legality and numerical admissibility, but the frozen surrogate-trained generation rule did not retain enough teacher-aligned ranking precision to reach the top competitive region at the largest native setting.

The bounded result is therefore informative rather than merely negative. It identifies where the current PPO formulation is useful and where it begins to lose reliability. In smaller native settings, the policy can concentrate a limited number of candidates into competitive regions of the teacher-labelled landscape. In larger settings, the same pipeline remains executable but loses rank concentration. This points to a clear route for future high-\(N\) improvement: policy generation should be coupled with iterative teacher-feedback updates, uncertainty-aware candidate acquisition and targeted Abaqus validation near surrogate-favoured but teacher-uncertain regions.

The broader implication for RL-for-AM is that sequence-generation algorithms should be evaluated by their ability to produce physically testable candidates under a limited validation budget, not only by whether they produce a single new record. For computationally expensive manufacturing optimisation, policy-level proposal generation is valuable when paired with explicit teacher validation and a clear claim boundary. The present results therefore support teacher-validated, small-\(N\)/top-\(k\) competitiveness and expose a high-\(N\) reliability boundary that future finite-element-grounded policy-generation methods must address.

\subsection{Physical interpretation: reward hierarchy and scale-separated scan-order tendencies}

The reward hierarchy provides a useful lens for interpreting the teacher-validated scan-order patterns, but the interpretation must remain bounded. The selected teacher-validated PPO-generated sequences do not prove a universal scan-order law. Rather, they suggest an interpretable ordering tendency under the present benchmark, material model, heat-input definition and Abaqus teacher formulation. Within this evidence boundary, the observed patterns can be read as a physically plausible compromise between global heat dispersion and local thermal continuity.

The connection to the reward hierarchy is direct. Section 3.2 showed that the terminal reward is governed by a physically gated ordering: \(U_2\) defines the primary warpage-admissibility layer, \(\mathrm{PEEQ}\) acts as the plasticity-safety filter, and \(\mathrm{Surface}_T\) provides conditional secondary ranking inside the admissible response region. This hierarchy discourages a sequence that improves a secondary stress-related response at the cost of excessive geometric distortion. It also prevents scan-order quality from being interpreted as a single-field optimisation problem. A competitive sequence must satisfy a constrained thermo-mechanical balance rather than minimise one response in isolation.

This helps explain why the selected scan orders do not simply reproduce a single classical scan rule. A raster or block strategy emphasises local continuity; an odd-even or regular-jump strategy emphasises separation; a centre-out strategy emphasises symmetry or regional expansion \cite{Dar2025,Ding2024}. Each heuristic encodes one dominant geometric intuition. The finite-element-grounded reward, by contrast, evaluates the completed thermo-mechanical history of the sequence. It can therefore favour orderings that mix several local and global principles when no single predefined rule is sufficient to satisfy the warpage, plasticity and secondary stress-response hierarchy simultaneously.

The representative teacher-validated scan orders suggest such a mixed structure. They do not appear as purely monotonic raster paths, purely random dispersion patterns or uniformly spaced long-jump sequences. Instead, the selected patterns show broad spatial dispersion across larger regions together with shorter-range structured grouping. In descriptive terms, this can be interpreted as a scale-separated scan-order tendency: global dispersion at the large scale, local structured grouping at the short range.

\begin{figure}[htbp]
  \centering
  \includegraphics[width=0.95\textwidth]{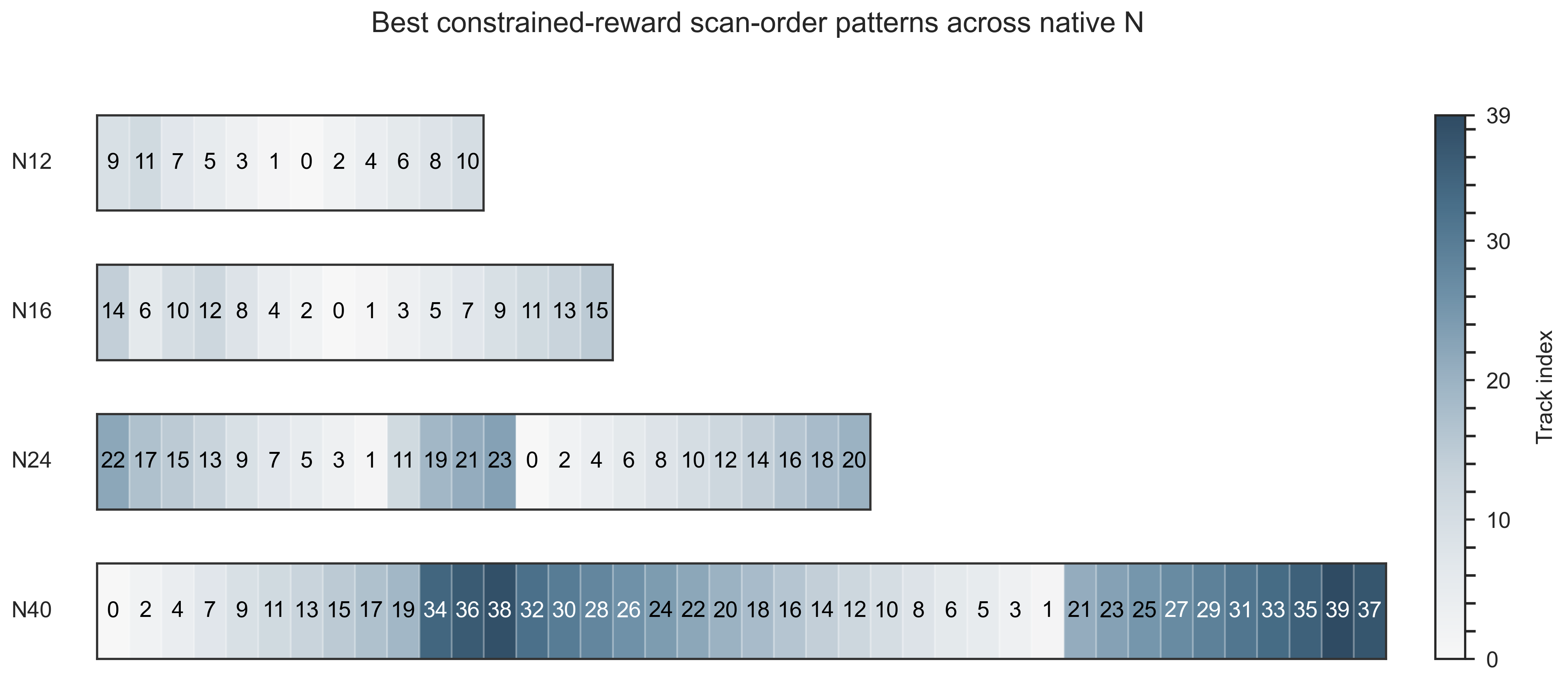}
  \caption{Representative teacher-validated scan-order tendencies. The selected teacher-validated scan orders illustrate an interpretable scale-separated pattern, with broad spatial dispersion combined with local structured grouping. The figure is used to support physical interpretation of the validated scan-order patterns, not to claim a universal scan rule.}
  \label{fig:4.2}
\end{figure}

A physically plausible explanation is that the two components address different length scales of the reward hierarchy, consistent with prior DED studies showing that deposition sequence, cumulative thermal history and return/dwell time affect distortion and residual-stress development \cite{Caiazzo2022,Denlinger2015,Ding2024}.

\begin{itemize}
  \item At the large scale, global dispersion increases the spatial separation between consecutive heat-input events across the broader deposition domain. This reduces the likelihood of sustained heat concentration in one region and is directly aligned with the \(U_2\)-first layer of the reward hierarchy, where warpage-related geometric admissibility defines the primary manufacturing constraint.

  \item At the short range, local structured grouping avoids the opposite extreme of fully fragmented long-distance jumping. By retaining ordered deposition within local neighbourhoods, the sequence can preserve a more continuous local thermal history and reduce abrupt transient thermal-gradient loading between neighbouring tracks. This short-range structure is consistent with the need to avoid unfavourable plasticity and secondary stress-response behaviour after the primary warpage condition has been considered.
\end{itemize}

The observed scale-separated tendency can therefore be interpreted as a constrained thermo-mechanical compromise: disperse heat broadly enough to suppress macroscopic distortion, but retain enough local order to avoid overly discontinuous reheating histories.

This interpretation also clarifies why the policy-generated patterns should not be judged only against single-principle human heuristics. A sequence that is globally well dispersed may still create unfavourable local thermal histories. Conversely, a locally continuous sequence may concentrate heat excessively at the regional scale. The reward hierarchy imposes both types of pressure. The resulting PPO-generated candidates are therefore better understood as finite-element-filtered compromises between competing thermal and mechanical requirements, rather than as simple variants of raster, odd-even or block scanning.

The interpretation remains conditional on the present evidence. The motif has not been quantified as a general design law, and the current PPO batch did not demonstrate all-N dominance. In particular, the high-N performance boundary reported in Section 4.2 shows that the presence of an interpretable ordering tendency does not guarantee top-k competitiveness at larger scan-order horizons. The motif should therefore be treated as a hypothesis-generating pattern emerging from the validated candidates, not as a universal prescription for LDED scan planning.

The value of this interpretation is that it makes the policy-generated scan orders physically interpretable to manufacturing and metallurgical researchers. In the present evidence setting, the framework does not merely return opaque permutations or select among named engineering rules; it can also generate intermediate ordering structures that combine multiple physical intuitions and can be interpreted through the finite-element response hierarchy. This provides a bridge between human-designed scan-strategy reasoning and high-dimensional policy-based sequence generation.

\subsection{From validated track-order tendencies to area-level priority-field planning}

The scale-separated scan-order tendency discussed in Section 4.3 suggests that a teacher-validated scan-order sequence contains more than an ordered list of track indices. It also contains spatial sequencing information: which regions are prioritised early, which regions are delayed, and how broad dispersion is balanced against local grouping. This observation motivates a conceptual extension from validated track-level ordering to area-level priority-field planning, consistent with broader region-, island- and path-pattern-based scan-planning ideas used to manage heat accumulation and residual-stress response in additive manufacturing \cite{Cheng2023,Ding2024}. This extension is not introduced as an additional validated result, but as a way to reuse the spatial ordering information extracted from teacher-validated scan-order candidates for future planning representations.

The evidence boundary of the present study remains one-dimensional track-order optimisation. Each validated candidate is represented as a permutation of deposition tracks, and all physical-performance claims are based on Abaqus-labelled N12, N16, N24 and N40 cases. Nevertheless, once a validated scan order has been obtained, it can be deterministically re-expressed as a spatial priority profile. For a scan-order permutation \(\sigma = (a_0, a_1, \ldots, a_{N-1})\), let \(r_\sigma(i)\) denote the position of track \(i\) in the sequence. A normalised priority score can then be assigned to each track:
\[
  s_\sigma(i) = \epsilon + (1 - 2\epsilon)\left(1 - \frac{r_\sigma(i)}{N-1}\right),
  \qquad 0 < \epsilon \ll 1.
\]
Earlier-deposited tracks receive higher priority values, whereas later-deposited tracks receive lower values. The small offset \(\epsilon\) avoids degenerate zero-priority endpoints and keeps the resulting field numerically non-zero for downstream use. The score is not an additional learned variable and does not alter the teacher-validated ranking. It is a deterministic transformation that converts a validated sequence into a reusable spatial planning prior.
The one-dimensional priority profile can then be lifted into a two-dimensional discrete field. For a virtual cell or patch indexed by \((i,j)\), one simple symmetric construction is a Euclidean lifting of the two one-dimensional priority components:
\[
  \tilde{s}_{\mathrm{2D}}(i,j) = \sqrt{s_\sigma(i)^2 + s_\sigma(j)^2}.
\]
This construction treats the two one-dimensional priority values as orthogonal spatial components and combines them through an \(L_2\)-norm. In this interpretation, a location in the two-dimensional field inherits priority from two independent sequencing directions, and the resulting value represents their combined priority magnitude rather than their average. If a bounded visual or numerical field is required, the unnormalised field can be rescaled after construction:
\[
  s_{\mathrm{2D}}(i,j) = \frac{\tilde{s}_{\mathrm{2D}}(i,j)}{\max_{p,q} \tilde{s}_{\mathrm{2D}}(p,q)}.
\]
The lifting rule is therefore used as a deterministic geometric transformation, not as an additional physical reward model. Other lifting rules, including arithmetic averaging, root-mean-square aggregation, tensor-product coupling or physics-informed kernels, could be adopted if a future area-level planner imposes additional thermal, geometric or machine-kinematic constraints.

This transformation formalises the following information pathway:
\[
  \begin{aligned}
    &\text{validated track-level order}
    \longrightarrow \text{rank-derived priority profile} \\
    &\longrightarrow \text{area-level scan-priority field}.
  \end{aligned}
\]

The priority field is therefore not a scan path. It does not enforce path continuity, travel-distance cost, local reheating avoidance, hatch-turn constraints, machine dynamics or explicit thermal accumulation. It also does not generate new Abaqus teacher labels or imply improved \(U_2\), \(\mathrm{PEEQ}\) or \(\mathrm{Surface}_T\). Any physical claim at the area level would require a separate path-generation rule followed by new finite-element teacher validation.

\begin{figure}[htbp]
  \centering
  \includegraphics[width=0.95\textwidth]{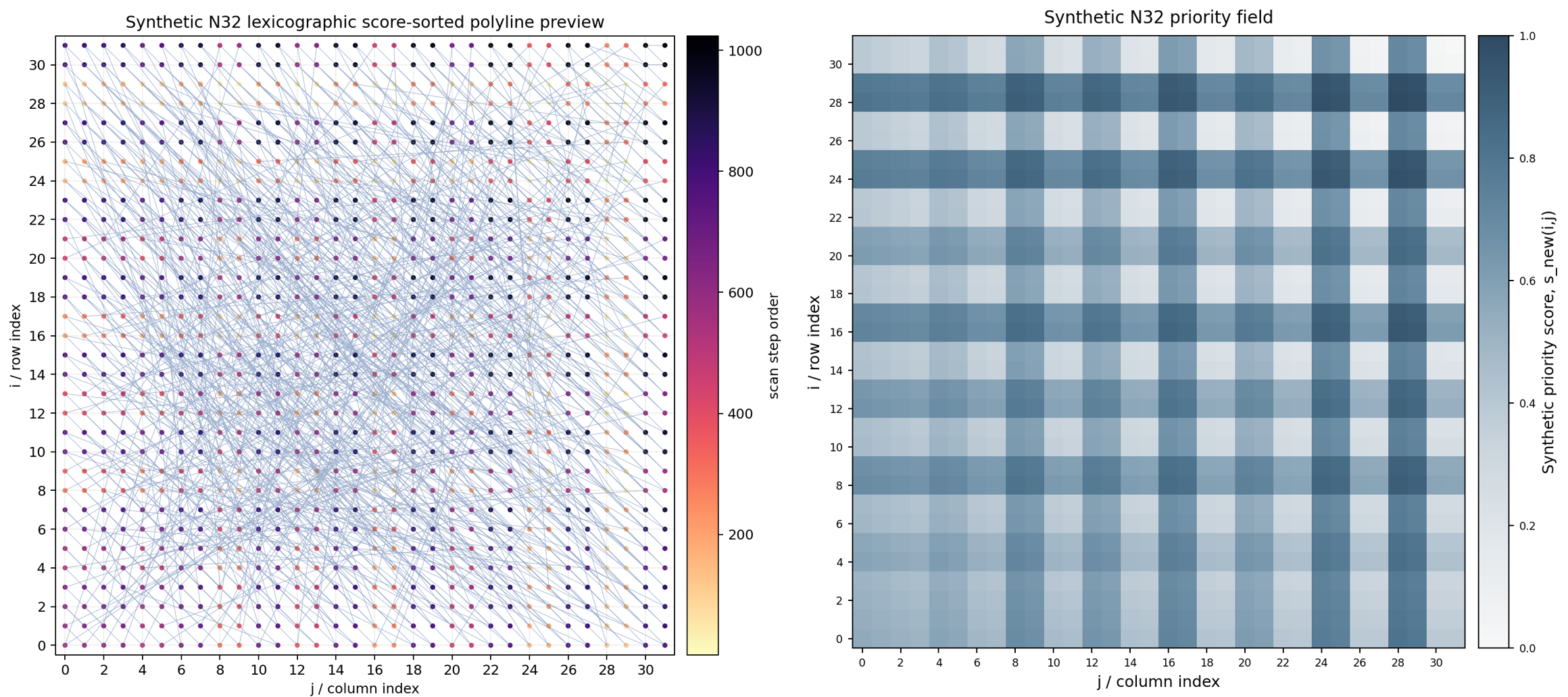}
  \caption{Conceptual extension from validated track-level scan order to an area-level scan-priority field. (a) Rank-derived priority structure obtained from a teacher-validated one-dimensional track-order candidate. (b) Corresponding \(M \times M\) scan-priority field \(s_{\mathrm{2D}}(i,j)\), with \(M=32\) used only as an illustrative discretisation. The field re-expresses track-level ordering information as a spatial planning prior and is not a validated two-dimensional scan path.}
  \label{fig:4.3}
\end{figure}

The value of this construction is information reuse. A validated one-dimensional scan order contains spatial sequencing information that would otherwise remain locked inside a permutation. The priority-field representation converts that information into a form that can seed future patch-, island- or cell-based scan planners \cite{Cheng2023,Ding2024}, while preserving the central logic of the present framework: learned policies may propose ordering priors, but physical validity remains contingent on teacher validation.

This provides a forward-looking bridge from finite-element-grounded track-order learning to area-level scan-path planning. It does not expand the performance claims of the present study beyond the validated native track-order cases. Instead, it shows how the interpretable ordering tendencies identified in Section 4.3 could be translated into a higher-resolution planning representation in which deposition priority is assigned across two-dimensional regions rather than along a fixed set of parallel tracks.

\subsection{Limitations, longer-horizon rollouts and validation outlook}

The present study establishes a finite-element-grounded PPO scan-order generation workflow, but its claim boundary remains deliberately limited. The policy was trained in a surrogate terminal-reward environment, whereas final performance assessment was made only after independent Abaqus validation. The policy was not trained online inside Abaqus, the surrogate was not treated as a replacement for the finite-element teacher, and the PPO-generated candidates did not establish new native-reference records or exceed the mature surrogate-assisted best records. The supported claim is therefore bounded: the framework generated legal, executable and teacher-validated scan-order candidates, with small-\(N\)/top-\(k\) competitiveness and a clear high-\(N\) reliability boundary.

The main limitation is the decline in ranking reliability as the scan-order horizon increases. The N40 candidates remained finite-element executable, but they did not enter the top-\(k\) competitive region of the native reference pool. This shows that the high-\(N\) limitation is not scan-order legality or numerical admissibility, but the ability of a frozen surrogate-trained policy to preserve teacher-aligned ranking precision in an expanding permutation space, a known challenge for learning-based combinatorial optimisation methods \cite{Bengio2021,Mazyavkina2021}.

Longer-horizon PPO rollouts, such as N50 or N64 sequences, can be useful as diagnostic probes of whether the scale-separated ordering tendency persists qualitatively beyond the validated native settings. However, without new Abaqus simulations, these rollouts remain policy outputs only. They should not be used to support claims about \(U_2\), \(\mathrm{PEEQ}\), \(\mathrm{Surface}_T\), residual stress or superiority over engineering baselines.

Future extensions should therefore couple policy generation with iterative teacher feedback. The surrogate-to-teacher alignment loss observed under policy-induced selection is consistent with optimisation-induced distribution shift: the policy preferentially samples high-predicted-reward regions that may be sparsely represented in the original teacher-labelled pool, reinforcing the need to treat the surrogate as a training and prioritisation model rather than as a final physical evaluator \cite{Kim2026,Francon2020}. A more robust validation loop would prioritise candidates that combine high predicted reward, high epistemic uncertainty and sufficient sequence diversity for Abaqus evaluation, following the broader logic of active surrogate improvement for expensive high-fidelity models \cite{Guo2024}. Newly labelled cases could then be added to the teacher-labelled pool to update the reward surrogate and refine the policy, rather than relying on uncalibrated out-of-distribution extrapolation.

The current validation is also limited to numerical finite-element teachers. Experimental LDED validation will be needed to test whether selected teacher-validated scan orders retain their advantages under real process variability, heat-loss uncertainty, material-model mismatch and microstructural effects. Similarly, the priority-field construction in Section 4.4 remains a planning prior rather than a validated two-dimensional toolpath. These limitations define the boundary of the present contribution: surrogate-trained PPO can generate physically admissible and teacher-validated scan-order candidates under a limited validation budget, while high-\(N\) reliability, experimental confirmation and area-level toolpath synthesis remain open validation targets.

\section{Conclusions}

This study developed a bilevel finite-element-grounded AI framework for scan-order optimisation in laser directed energy deposition. The scan-order problem was formulated as a finite-horizon sequential decision process, where each action selects the next unprocessed track and each episode defines a complete scan-order permutation. A supervised surrogate terminal-reward environment was trained from native Abaqus teacher-labelled data, and a frozen MaskablePPO policy was used to generate scan-order candidates for independent finite-element validation. Across the native N12, N16, N24 and N40 settings, the PPO-generated candidates were admitted to Abaqus thermo-mechanical simulation and terminal teacher-metric extraction. The evidence supports a bounded conclusion: surrogate-trained PPO can generate legal, executable and teacher-validated scan-order candidates, with small-\(N\)/top-\(k\) competitiveness and a clear high-\(N\) reliability boundary. The PPO layer did not establish new native-reference records and is not claimed to exceed the mature surrogate-assisted best records.

The study makes three main contributions.

\begin{enumerate}
  \item \textbf{A bilevel AI evidence chain combining surrogate-assisted teacher-guided optimisation with PPO policy generation.} The surrogate-assisted loop learns the Abaqus-labelled response landscape and provides the strongest current search mechanism within the finite-element teacher-labelled evidence pool. The MaskablePPO layer then distils this learned terminal-reward landscape into an executable sequential decision policy capable of autonomously generating legal scan-order candidates. The PPO-generated candidates are not accepted on the basis of surrogate score alone; their physical performance is determined by independent Abaqus teacher validation and teacher-metric ranking. This establishes PPO as a teacher-validated policy-generation layer rather than as a standalone record-level optimiser.

  \item \textbf{A physically gated reward hierarchy for LDED scan-order optimisation.} The teacher-labelled response landscape shows that scan-order quality should not be reduced to a flat weighted compromise among distortion, plasticity and stress-related metrics. Instead, the results support a physically ordered hierarchy: \(U_2\) first defines the warpage-admissible geometric window, \(\mathrm{PEEQ}\) acts as a plasticity-safety filter, and \(\mathrm{Surface}_T\) provides conditional secondary ranking inside the admissible response region. This hierarchy reframes reward design as a constrained manufacturing decision: preserve geometric admissibility first, then exploit the remaining response window for residual-stress-related improvement.

  \item \textbf{An interpretable scan-order tendency beyond single-rule human heuristics.} The teacher-validated PPO candidates do not simply reproduce raster, block, odd-even or uniform long-jump strategies. Their selected patterns suggest a scale-separated ordering tendency: global dispersion across broad regions combined with local structured grouping over shorter ranges. This indicates that effective scan-order design in the present benchmark is neither pure spatial separation nor pure local continuity, but a constrained thermo-mechanical compromise between the two. The policy-generated sequences are therefore interpretable as process-conditioned ordering structures rather than opaque permutations alone.
\end{enumerate}

These contributions remain computationally bounded. The present study does not claim global optimality, exhaustive scan-order search, online real-time LDED control, arbitrary-\(N\) generalisation, dominance over conventional or learned scan-order strategies, superiority over the mature surrogate-assisted optimiser, validated N50/N64 performance or a validated two-dimensional area scan path. The N50/N64 candidates remain longer-horizon rollout diagnostics, and the two-dimensional priority-field formulation remains a planning prior for future scan-path synthesis. To our knowledge, this work represents one of the first finite-element-teacher-labelled reinforcement-learning attempts for scan-order optimisation in laser directed energy deposition. Its significance lies in moving RL-for-LDED from surrogate-only sequence scoring toward teacher-validated scan-order policy generation and knowledge discovery.

Three future directions follow directly from the present results.

\begin{enumerate}
  \item \textbf{Closed-loop teacher-feedback learning.} Future policy generation should be coupled to iterative Abaqus feedback. Candidates with high predicted reward, high epistemic uncertainty and sufficient sequence diversity should be prioritised for teacher validation, added to the labelled pool and used to update both the reward surrogate and the policy. This would move the framework from frozen surrogate-guided generation toward active finite-element-grounded learning.

  \item \textbf{Experimental ranking transferability.} The next experimental step is to test whether Abaqus-ranked scan orders remain favourable under physical LDED measurements. The key validation target should be ranking transferability, not exact point-wise numerical agreement: whether teacher-preferred candidates retain advantages in measured distortion, residual-stress response, melt-track quality and microstructural variation.

  \item \textbf{Area-level toolpath synthesis.} The validated track-order tendencies and priority-field construction provide a route from one-dimensional sequence optimisation to two-dimensional scan-planning priors. Future work should convert these fields into executable area-level toolpaths by adding path-continuity, travel-distance, local reheating, hatch-transition, machine-kinematic and thermal-accumulation constraints, followed by new finite-element teacher validation.
\end{enumerate}

Within these boundaries, the study establishes a computationally validated RL--FEA framework for offline LDED scan-order optimisation. More broadly, it shows that under physically grounded simulation feedback, learning-based policies can do more than fit process data or select from existing scan rules. They can propose candidate scan-order structures, subject them to teacher validation and support physical interpretation, providing a foundation for machine-discovered process knowledge in laser additive manufacturing.

\section*{Author Contributions}

\textbf{Xian Wu}: Conceptualization, methodology, software, finite-element modelling, reinforcement-learning framework implementation, formal analysis, investigation, data curation, visualization, and writing -- original draft.

\textbf{Haoran Li}: Methodology, reinforcement-learning discussion, and writing -- review and editing.

\textbf{Yuanqi Chu}: Methodology, mathematical modelling, formal analysis, and writing -- review and editing.

\textbf{Dongbin Zhao}: Supervision, reinforcement-learning methodology guidance, and writing -- review and editing.

\textbf{Bin Wang}: Supervision, project administration, research guidance, and writing -- review and editing.

All authors reviewed and approved the final manuscript.

\section*{Declaration of Competing Interest}

The authors declare that they have no known competing financial interests or personal relationships that could have appeared to influence the work reported in this paper.

\section*{Acknowledgments}

The authors acknowledge Brunel University London for providing the research environment and computational/software resources that supported this work. The authors also acknowledge the use of Abaqus-based finite-element modelling and post-processing in the simulation-based reference evaluation workflow.

\section*{Data availability}

The data and code supporting the findings of this study are available in the project GitHub repository:

\url{https://github.com/BrunelXian/RL-LAM-ScanOpt}

The repository includes scripts, processed tables, figure-generation outputs, and diagnostic analysis assets used for proxy evaluation, FEA reference-label analysis, ranking robustness checks, and Proxy--FEA alignment. Large binary simulation files, including Abaqus ODB files, are not included because of file-size and software-specific storage constraints. The derived metrics and processed outputs required to reproduce the reported tables and figures are provided.

\bibliographystyle{elsarticle-num}
\bibliography{references}

\end{document}